\documentclass[sigconf,nonacm]{acmart}

\AtBeginDocument{%
  \providecommand\BibTeX{{%
    \normalfont B\kern-0.5em{\scshape i\kern-0.25em b}\kern-0.8em\TeX}}}

\setcopyright{acmcopyright}
\copyrightyear{2024}
\acmYear{2024}
\acmDOI{XXXXXXX.XXXXXXX}
\usepackage{multirow}
\usepackage{subcaption}
\usepackage{xspace}
\usepackage{amsfonts}
\usepackage{amsbsy}
\usepackage{graphicx}
\usepackage{color,soul}

\acmISBN{978-1-4503-XXXX-X/18/06}

\newcommand{\name}{\texttt{LimitNet}\xspace}
\newcommand{\etal}{\textit{et al.}\xspace}
\newcommand{\specialcell}[2][c]{%
  \begin{tabular}[#1]{@{}c@{}}#2\end{tabular}}

\newcommand{\note}[1]{{\color{red}{#1}}}
\renewcommand{\note}[1]{#1}
\newcommand{\review}[1]{{\color{red}{#1}}}
\renewcommand{\review}[1]{#1}

\begin{document}

\acmYear{2024}\copyrightyear{2024}
\acmConference[MOBISYS '24]{The 22nd Annual International Conference on Mobile Systems, Applications and Services}{June 3--7, 2024}{Minato-ku, Tokyo, Japan}
\acmBooktitle{The 22nd Annual International Conference on Mobile Systems, Applications and Services (MOBISYS '24), June 3--7, 2024, Minato-ku, Tokyo, Japan}
\acmDOI{10.1145/3643832.3661856}
\acmISBN{979-8-4007-0581-6/24/06}

%%
%% The "title" command has an optional parameter,
%% allowing the author to define a "short title" to be used in page headers.
% \title{\name: Content-Aware Lightweight Progressive Image AutoEncoder for Embedded Devices under Limited Dynamic Network Bandwidth}

\title{\name: Progressive, Content-Aware Image Offloading for Extremely Weak Devices \& Networks}

%%
%% The "author" command and its associated commands are used to define
%% the authors and their affiliations.
%% Of note is the shared affiliation of the first two authors, and the
%% "authornote" and "authornotemark" commands
%% used to denote shared contribution to the research.
\author{Ali Hojjat}
\affiliation{%
  \institution{Kiel University}
  \country{Germany}}
\email{aho@informatik.uni-kiel.de}

\author{Janek Haberer}
\affiliation{%
  \institution{Kiel University}
  \country{Germany}}
\email{jha@informatik.uni-kiel.de}

\author{Tayyaba Zainab}
\affiliation{%
  \institution{Kiel University}
  \country{Germany}}
\email{tza@informatik.uni-kiel.de}

\author{Olaf Landsiedel}
\affiliation{%
  \institution{Kiel University}
  \country{Germany}}
\email{ol@informatik.uni-kiel.de}

%%
%% By default, the full list of authors will be used in the page
%% headers. Often, this list is too long, and will overlap
%% other information printed in the page headers. This command allows
%% the author to define a more concise list
%% of authors' names for this purpose.

%%
%% The abstract is a short summary of the work to be presented in the
%% article.
\begin{abstract}
IoT devices have limited hardware capabilities and are often deployed in remote areas. Consequently, advanced vision models surpass such devices' processing and storage capabilities, requiring offloading of such tasks to the cloud. However, remote areas often rely on LPWANs technology with \textit{limited bandwidth, high packet loss rates}, and \textit{extremely low duty cycles}, which makes fast offloading for time-sensitive inference challenging. 
Today's approaches, which are deployable on weak devices, generate a non-progressive bit stream, and therefore, their decoding quality suffers strongly when data is only partially available on the cloud at a deadline due to limited bandwidth or packet losses.

In this paper, we introduce \name, a progressive, content-aware image compression model designed for extremely weak devices and networks. \name's lightweight progressive encoder prioritizes critical data during transmission based on the content of the image, which gives the cloud the opportunity to run inference even with partial data availability.

Experimental results demonstrate that \name, on average, compared to SOTA, achieves 14.01 p.p. (percentage point) higher accuracy on ImageNet1000, 18.01 pp on CIFAR100, and 0.1 higher mAP@0.5 on COCO. Also, on average, \name saves 61.24\% bandwidth on ImageNet1000, 83.68\% on CIFAR100, and 42.25\% on the COCO dataset compared to SOTA, while it only has 4\% more encoding time compared to JPEG (with a fixed quality) on STM32F7 (Cortex-M7).

\end{abstract}

%%
%% The code below is generated by the tool at http://dl.acm.org/ccs.cfm.
%% Please copy and paste the code instead of the example below.
%%
\begin{CCSXML}
<ccs2012>
   <concept>
       <concept_id>10010147.10010257</concept_id>
       <concept_desc>Computing methodologies~Machine learning</concept_desc>
       <concept_significance>500</concept_significance>
       </concept>
   <concept>
       <concept_id>10010520.10010553</concept_id>
       <concept_desc>Computer systems organization~Embedded and cyber-physical systems</concept_desc>
       <concept_significance>300</concept_significance>
       </concept>
   <concept>
       <concept_id>10003120.10003138</concept_id>
       <concept_desc>Human-centered computing~Ubiquitous and mobile computing</concept_desc>
       <concept_significance>300</concept_significance>
       </concept>
 </ccs2012>
\end{CCSXML}

\ccsdesc[500]{Computing methodologies~Machine learning}
\ccsdesc[300]{Computer systems organization~Embedded and cyber-physical systems}
\ccsdesc[300]{Human-centered computing~Ubiquitous and mobile computing}

\keywords{Deep Learning, Edge Computing, Lightweight AutoEncoders, Content-Aware Encoding, Image Compression, Progressive Offloading, Progressive Compression, Internet of Things}

\maketitle

\baselineskip=12bp

\section{Introduction}

\begin{figure}[ht]
  \centering
   \includegraphics[trim=260 80 210 110, clip, width=.5\textwidth]{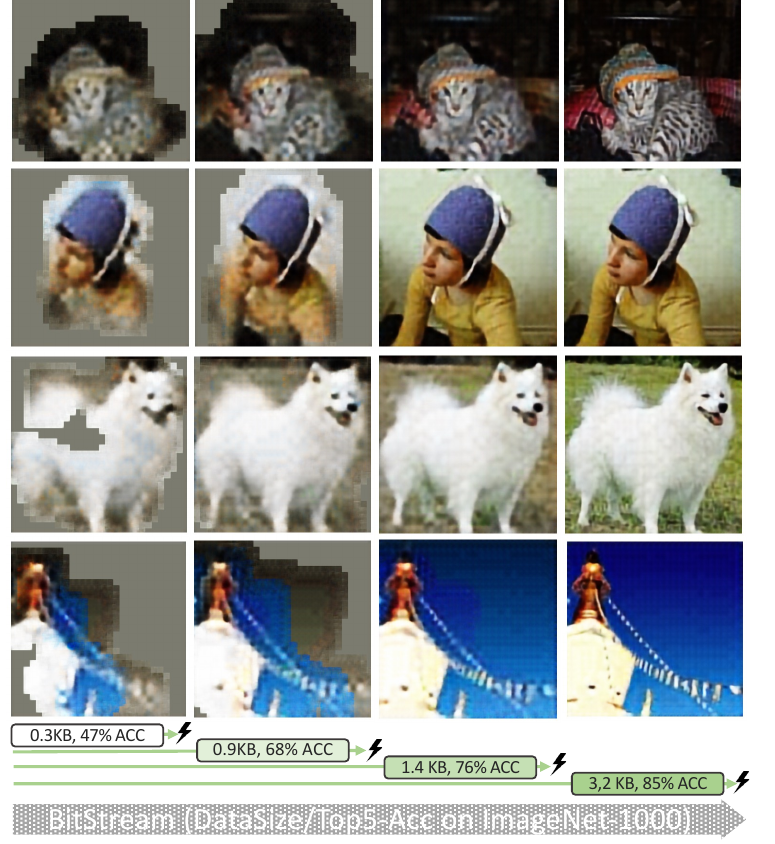}
  \caption{\review{Qualitative comparison of \name's progressive reconstruction. \name detects the important parts of the image and sends the encoded data in the order of importance.
  Progressive offloading allows the cloud to run the inference at any point (lightning symbol).
  The horizontal axis shows the offloading bitstream of an image, the first number shows the size of received data (in KB) at a specific time, followed by the corresponding Top5 Accuracy (Top5-Acc) of EfficientNet-B0 \cite{li2020efficient} on ImageNet1000 \cite{deng2009imagenet} using this received data.}}
   \label{fig:demo}
   \vspace{-1.5em}
\end{figure}

\noindent
\textbf{Motivation:} 
Neural networks enable new computer vision applications, such as classification and object detection, in the Internet of Things (IoT). For example, AI-enabled \review{distributed} IoT cameras enable emergency systems such as Apple's emergency SOS \cite{apple_emergency_sos} disaster response \cite{zainab2023lighteq, lim2022leqnet}, intruder detection \cite{vijayan2022video}, fire detection \cite{avazov2023forest}, \review{wild life monitoring \cite{camal2020building}} and road security surveillance \cite{ramachandran2021review}.
In such scenarios, it is crucial to have the classification result 
%on the cloud 
within a specific \textbf{deadline} to ensure a timely response \cite{choi2019latency} such as triggering alerts.

\begin{table*}[tb]
\caption{\note{Comparison of compression-offloading methods. "Offloading Granularity" measures the scale of the individual transmitted segments, and "Incomplete Data Accuracy" refers to the accuracy of the model when only a portion of data is available for decoding on the cloud side.}}
\centering
\begin{tabular}{|c|c|c|c|c|c|c|c|}
\hline
\textbf{Model}& \textbf{\specialcell{Offloading\\Granularity}} & \textbf{\specialcell{Transmission\\Cost}} & \textbf{\specialcell{Incomplete Data\\Accuracy}} & \textbf{\specialcell{Local\\Cost}}& \textbf{Objectives} \\

\hline
DeepCOD \cite{yao2020deep}&-& Med. & Low  & Low & Accuracy\\
BottleNet++ \cite{shao2020bottlenet++}&-&  High. & Low  & Low& Accuracy\\
AgileNN \cite{huang2022real}& - &  Med. & Low   & Low & Accuracy\\
SPINN \cite{laskaridis2020spinn}& Filter & Med. & Low  & Low& Accuracy\\
FLEET \cite{rethinking}& Set of filters & Med. & High  & Low& Accuracy\\
DynO \cite{almeida2022dyno}& - &  Low & Low & High& Accuracy\\
\hline
Starfish \cite{hu2020starfish}& - &  Low & Med. & Med. & Accuracy, Perception\\
JPEG \cite{wallace1992jpeg}& - & Low  & Low  & Low &Perception\\
ProgJPEG \cite{wallace1992jpeg}& DCT Scan & Low & Med. & Low &Perception\\
Ballé \etal \cite{balle2018variational}&- &  Low & Low  & High&Perception\\
Full-Res \cite{toderici2017full}&Latent&  Low & Med.   & High &Perception\\
DCCOI \cite{chen2021deep}&-& Med. & Low  & Low & Accuracy\\\
AccelIR \cite{ye2023accelir}&-&  Low & Low  & High &Accuracy\\
\hline

\textbf{\name}& \pmb{Subfilter} & \pmb{Low} & \pmb{High} & \pmb{Low}& \pmb{Accuracy}\\
\hline
\end{tabular}
\label{tab:model_comparison}
% \vspace{-1.5em}
\end{table*}

\noindent
\textbf{Challenges:} 
%In real-world scenarios, 
IoT devices are often deployed in remote areas and consist of embedded devices with \textbf{minimal power and hardware capabilities} to limit costs. 
As a result, modern vision models \cite{he2016deep, dosovitskiy2020image, redmon2016yolo} quickly exceed embedded systems' computing capabilities. Therefore, the input image must be compressed first \cite{hu2020starfish, yao2020deep, shao2020bottlenet++, wallace1991jpeg}, and then sent to the cloud for further processing. 
However, remote areas commonly have limited internet access, i.e., they often do not have access to cellular networking. In such settings, communication is usually limited to LPWAN technology \cite{chaudhari2020lpwan} (e.g., Sigfox \cite{lavric2019sigfox} and LoRa \cite{bor2016lora}) with \textbf{very low and often dynamic bandwidth} (less than 50 KB/s) as satellite communication is often too costly and energy-intensive \cite{chaudhari2020lpwan, fourati2021artificial}. 
\review{In these scenarios, IoT cameras use shared communication links, forcing them to have \textbf{limited duty cycles} (less than 1\%).}
In addition to these limitations, LPWANs have a \textbf{high packet loss rate}, which increases the transmission time by forcing multiple retransmissions \cite{sikora2019test}. Therefore, LPWAN cannot guarantee to transmit all data to the cloud within a predefined timeframe, e.g., a deadline or a given duty cycle budget.

To overcome these challenges, the system must be able to run inference on the cloud even with \textit{incomplete data}.

% \begin{figure}[tb]
%   \centering
%    \includegraphics[trim=5 5 5 5, clip, width=.45\textwidth]{Pics/Complexity.pdf}
%   \caption{Number of parameters vs. Top5 Accuracy of \name and established progressive image compression models \cite{toderici2017full, lee2022dpict, Hojjat_2023_CVPR, jeon2023context, johnston2018improved, zhang2019residual} on ImageNet1000 dataset \cite{deng2009imagenet} by using EfficientNet-B0 \cite{li2020efficient} as the classifier. The dashed line shows the classifier accuracy when we feed it the raw uncompressed input image.}
%    \label{fig:Complexity}
%    % \vspace{-1.5em}
% \end{figure}

\noindent
\textbf{Approach:}
When it comes to dealing with incomplete data, standard compression-offloading methods are not a solution since they produce an \textit{atomic latent}, which necessitates having all data for decoding. \textbf{Progressive Encoding}, known as \textit{fine-grained scalability (FGS)} \cite{li2001overview, said1996new}, is the common solution that enables a stepwise transmission of data, starting from low-resolution or coarse-grained information and gradually improving the quality or level of detail during transmission.  
Progressive encoding helps to prioritize critical data, optimize bandwidth usage, and, as a result, ensures that partial data, i.e., if the transmitter cannot transmit the complete compressed image before the deadline, can be efficiently used on the cloud side. 
Progressive compression \cite{balle2015density, balle2018variational, minnen2018joint,toderici2017full, Hojjat_2023_CVPR}, however, imposes a significant computational load due to its extensive parameterization, often comprising millions of parameters. To address this, creating a lightweight encoder -- as done in this work --  becomes essential, especially for weaker devices with hardware limitations.

\begin{figure*}[t!]
  \centering
   \includegraphics[trim=30 190 30 220, clip, width=1.0\textwidth]{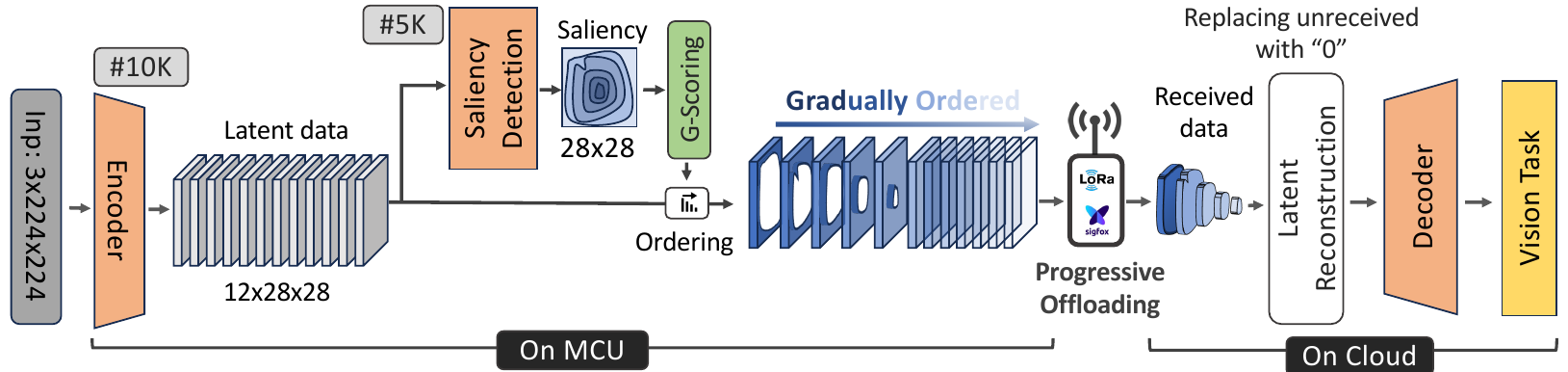}
   \caption{Overview of \name. Our encoder compresses the input to a latent representation. Gradual Scoring and our saliency detector extract the important parts of the latent data and assign an importance score to each latent data point. Afterward, \name starts to transmit the latent to the cloud in order of its importance score. On the cloud side, at any given time, we can reconstruct the latent by setting unreceived values to zero and feeding it to a powerful decoder. After reconstruction, we feed the output to a vision model \cite{redmon2016yolo, li2020efficient}.}
   \label{fig:overview}
\end{figure*}

\noindent

\note{\textbf{\name: } In this paper, we introduce \name, a lightweight, progressive, and content-aware image compression-offloading model designed for weak devices like Cortex-M33 or M7 series MCUs under extremely limited networks such as LPWANs.
In \name, we first design a lightweight CNN to encode the input into a latent representation.
Next, by designing a lightweight saliency detector network, we identify the important parts of the input image. However, using the saliency map out of the box leads to poor accuracy. To effectively use the saliency map, we introduce a "Gradual Scoring" algorithm, which enables the model to \textit{learn and choose} how much background, i.e., context, it needs during the training for better classification results; see Fig.~\ref{fig:demo}.
The output of Gradual Scoring is an importance score for each latent data point, indicating their relative impact on classification results.
Afterward, in order of the scores, we progressively transmit the data to the cloud. At any given time, the cloud can reconstruct the image and run the inference, which is essential when we have a deadline and deal with limited bandwidth. Also, in the case of packet loss, \name's prioritized approach ensures that the cloud side receives the most crucial parts for accurate classification.}

\note{As highlighted in Table~\ref{tab:model_comparison}, in comparison to the state of the art, \name 
%stands out as 
is the only method characterized by both low transmission and low local computational costs. Its progressiveness operates at a granularity level as small as a subfilter, achieving notably high accuracy even when only a portion of the encoded data is accessible on the cloud, as we show in our evaluation in Section \ref{Evaluation}}.

\noindent
\textbf{Contributions:}
\begin{enumerate}
\itemsep0em
\item \textit{\name} is a \textit{progressive content-aware encoder} incorporating a lightweight saliency detector and a novel Gradual Scoring mechanism. It is highly suitable for LPWANs, given their limited and dynamic network bandwidth, restricted duty cycles, and high packet loss rate (Sections~\ref{Saliency Detection},~\ref{Gradual Scoring},~\ref{Compression Efficiency} and~\ref{System-Level Benchmarks}).

\item \name is a {\textit{very lightweight image encoder}} with 15K parameters, efficiently executable on extremely weak devices, such as ARM Cortex-M series (Section~\ref{LightweightEncoder}).

\item We evaluate the performance of \name on ARM Cortex-M33 and M7 for vision tasks such as classification and object detection and evaluate system-level metrics such as RAM, Flash, current usage, execution time, and power consumption. We also evaluate \name's performance under different network bandwidths and different packet loss rates by comparing it with SOTA models (Sections~\ref{Compression Efficiency},~\ref{Evaluating Detail} and~\ref{System-Level Benchmarks}).

\end{enumerate}

% \newpage
\noindent
\textbf{Summary of results:}
\begin{enumerate}
\itemsep0em
\item \textit{Accuracy/mAP results}: For a given data size, \name on average achieves up to 14.01 p.p. (percentage point) higher accuracy compared to the SOTA on the ImageNet1000 dataset, 18.01 p.p. accuracy improvement on the CIFAR100 dataset, and 0.1 mAP@0.5 improvement on the COCO dataset.

\item \textit{Rate results}: For a given accuracy or mAP@50, \name on average saves 61.24\% bandwidth compared to the SOTA on the ImageNet1000 dataset, 83.68\% on the CIFAR100 dataset, and 42.25\% on the COCO dataset.

\item \textit{System benchmarks}: We deploy our model on two microcontrollers, nRF5340 (Cortex-M33) and STM32F7 (Cortex-M7). Our results show that \name needs 260 ms, 360 KB (17\%) RAM, and 107 KB (5\%) Flash to run on STM32F7 and 2189 ms, 344 KB (33\%) RAM, and 102 KB (10\%) Flash to run on nRF5340. \name only uses 4\% and 11\% more encoding time compared to JPEG (with a fixed quality) on STM and nRF, respectively, while, unlike JPEG, producing a progressive bitstream.

\end{enumerate}

\section{Background and Motivation}

This section presents the required background, limitations, and challenges of existing compression models and LPWANs.

\subsection{Compression models}
\subsubsection{Stand-alone image compression models}
We categorize image compression models into two groups: classical approaches such as JPEG \cite{wallace1992jpeg}, JPEG2000 \cite{skodras2001jpeg}, WebP\cite{webp}, BPG \cite{bpg}, and more recent ones which use the power of deep learning for compression, including Hyperprior \cite{balle2018variational}, Full-Resolution \cite{toderici2017full} and other methods \cite{minnen2018joint, cheng2020learned, su2020scalable, Balle2016end, ye2023accelir}. While recent deep compression models exhibit excellent performance, they cannot be utilized on embedded MCUs due to their stark resource demands.
Starfish \cite{hu2020starfish} addresses these constraints by introducing a lightweight image encoder. Further, it incorporates dropout in the bottleneck to account for the network's unreliability. To the best of our knowledge, Starfish is the only deep image compression approach that is executable on the MCUs and addresses the dynamic nature of wireless networks.
In addition to having a light encoder, we argue that it is crucial to prioritize important data when it comes to limited bandwidths, which Starfish cannot do since it has a content-agnostic encoder. In contrast, \name features a lightweight, progressive, content-aware encoder, which prioritizes the important data during the offloading. 
As a result, it ensures that data arrives in the order of importance, allowing the cloud to gracefully run inference even with partial data availability.

\noindent
\subsubsection{Compression for offloading}
In resource-constrained networks, offloading data to a more computationally capable node is a potential solution to overcome resource challenges on these devices. Besides image compression models, numerous works focus on data compression in offloading settings, i.e., compressing the data before offloading it to the cloud for further processing. In general, we can group the offloading techniques into two categories \cite{matsubara2022split}: offloading with autoencoders \cite{matsubara2019distilled, hu2020fast, jankowski2020joint, choi2020back, matsubara2020head, eshratifar2019bottlenet} and without autoencoder \cite{zeng2019boomerang, pagliari2020crime, li2018learning, li2018auto, jeong2018computation, eshratifar2019jointdnn}. The autoencoder-based models such as DeepCOD \cite{yao2020deep} and BottleNet++ \cite{shao2020bottlenet++} are more suitable for the situation where we have limited network bandwidth since they compress the data before offloading.

Despite their advantages, these offloading models produce a non-progressive content agnostic bitstream, which needs all data for decoding and does not consider the effect of each data point in the target task. In contrast, \name features a progressive, content-aware encoder, which prioritizes the important data during the offloading.
It ensures that data arrives in the order of importance, allowing the cloud to gracefully run inference even with partial data availability

\subsection{LPWAN} \label{LPWAN}
LPWANs, including Sigfox, LoRaWAN, and NB-IoT, are wireless communication technologies specifically designed to facilitate long-range, low-power connectivity in the IoT field. 
In contrast to cellular technologies and WiFi, LPWANs utilize lower frequencies to efficiently transmit data over extended distances: Sigfox covers up to 40km, LoRaWAN up to 20km, and NB-IoT up to 10km. However, LPWANs provide very limited bandwidth: Sigfox offers 100 B/s with a 1\% duty cycle, LoRaWAN provides 0.3 KB/s to 50 KB/s also with a 1\% duty cycle, and NB-IoT delivers up to 200 KB/s \cite{bor2016lora, lavric2019sigfox, chaudhari2020lpwan}.
For example, sending a 10 KB image using Sigfox takes roughly 800 seconds at 100 B/s. Similarly, with LoRaWAN, at 10 KB/s, the process takes about 8 seconds. However, due to the 1\% duty cycle, a node can only use the network for 36 seconds per hour on average, which becomes quickly noticeable when, for example, dealing with multiple image transmissions in time-sensitive scenarios.
Further, due to wireless link dynamics, connection quality might unexpectedly deteriorate, forcing the LPWAN to retransmit data and potentially even change to a more robust encoding, which both, in turn, provide less bandwidth and thereby increase transmission times and the radio duty cycle \cite{bor2016lora, sikora2019test}. 
Therefore, LPWAN cannot guarantee to transmit all data within a timeframe, e.g., a deadline or a given duty cycle budget.
Thus, it is essential to be able to operate on partial data.

\name addresses these issues by incorporating a \textit{progressive encoder} that identifies important parts of the latent space for classification accuracy. 
By sending data in order of importance and retransmitting it as needed until the time budget for transmission, i.e., deadline or duty cycle, is reached, \name ensures that the most important data is available at the cloud, and it gracefully produces an output image from partial data.

\begin{figure}[tb]
  \centering
   \includegraphics[trim=237 253 165 200, clip, width=.49\textwidth]{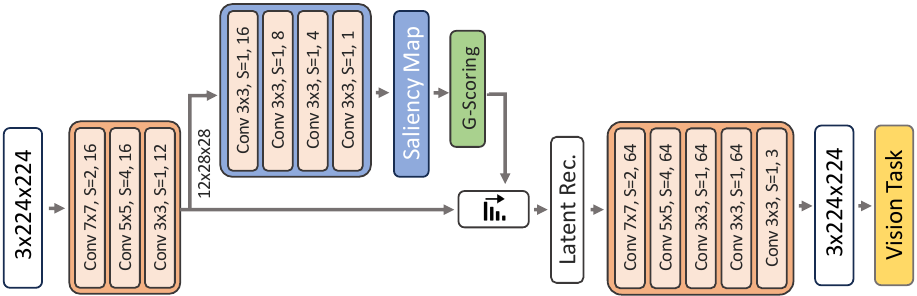}
   \caption{The architecture of \name is inspired by ResNet \cite{he2016deep} and adopts larger kernels for the first layers. All the used layers are supported by the DSP accelerator of the MCUs.}
   \label{fig:arch}
   % \vspace{-1em}
\end{figure}

\section{\name}
We begin this section by presenting the key components of \name before detailing its architecture, training phases, offloading mechanism, and quantization process.

\subsection{Overview} \label{Overview}

\begin{figure*}[tb]
  \centering
   \includegraphics[trim=20 190 33  217, clip, width=1.0 \textwidth]{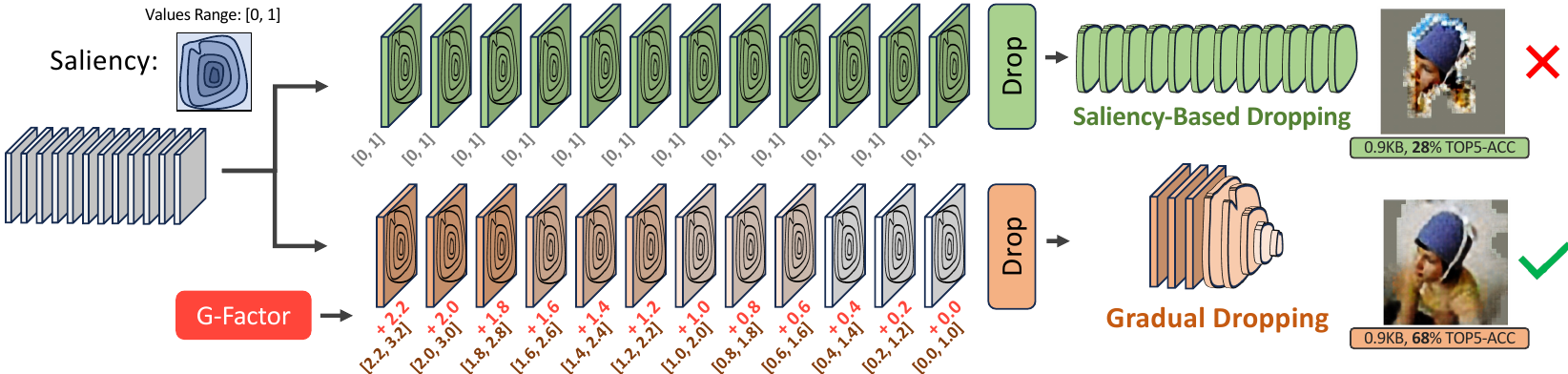}
   \caption{\note{Details of Gradual Scoring. This Figure illustrates where and how we add $G_{Factor}$ to each filter's activations of the latent, enabling the model to learn and choose how much background, i.e., context, it needs.}}
   \label{fig:scoring}
\end{figure*}

\note{In \name, we first encode the input image to a latent space utilizing a lightweight encoder; see Fig.~\ref{fig:overview}. Then, with a lightweight saliency detection branch, we detect the important parts of the input image. However, using this saliency map directly on the latent representation causes bias to the foreground, resulting in blocky outputs and subpar accuracy, see Fig.~\ref{fig:scoring}.
To address this, we introduce a "Gradual Scoring" algorithm, which enables the model to learn and determine the necessary background dynamically.
The output of Gradual Scoring is an importance score for each data point in the latent, indicating their relative impact on the classification accuracy.
Afterward, based on these scores, we progressively transmit the data to the cloud, where we run the image decoder and the classifier. The cloud can reconstruct the image and gracefully run inference at any time. In the case of packet loss, \name's prioritized approach ensures that the cloud side receives the most important data first.}

\subsection{Lightweight Encoder}  \label{LightweightEncoder}
We design an asymmetric autoencoder \cite{kramer1991nonlinear, yao2020deep} with a lightweight encoder ${ENC_{\theta_{ENC}}}$ for the edge and a deeper decoder ${DEC_{\theta_{DEC}}}$ for the cloud. The encoder gets the input image $X^{C \times H \times W}$ and compresses it to $Z^{L \times K \times K}$:

\begin{equation}
X^{C \times H \times W} \rightarrow {ENC_{\theta_{ENC}}}(X^{C \times H \times W}) \rightarrow Z^{L \times K \times K}
  \label{eq:1}
\end{equation}

Where ${C \times H \times W}$ shows the input size, and ${L \times K \times K}$ shows the shape of the latent data. Inspired by ResNet \cite{he2016deep}, we use large kernels (3, 5, and 7) for the first layers; see Fig.~\ref{fig:arch}.

\begin{figure}[tb]
  \centering
   \includegraphics[trim=50 105 100 99, clip, width=.57\textwidth]{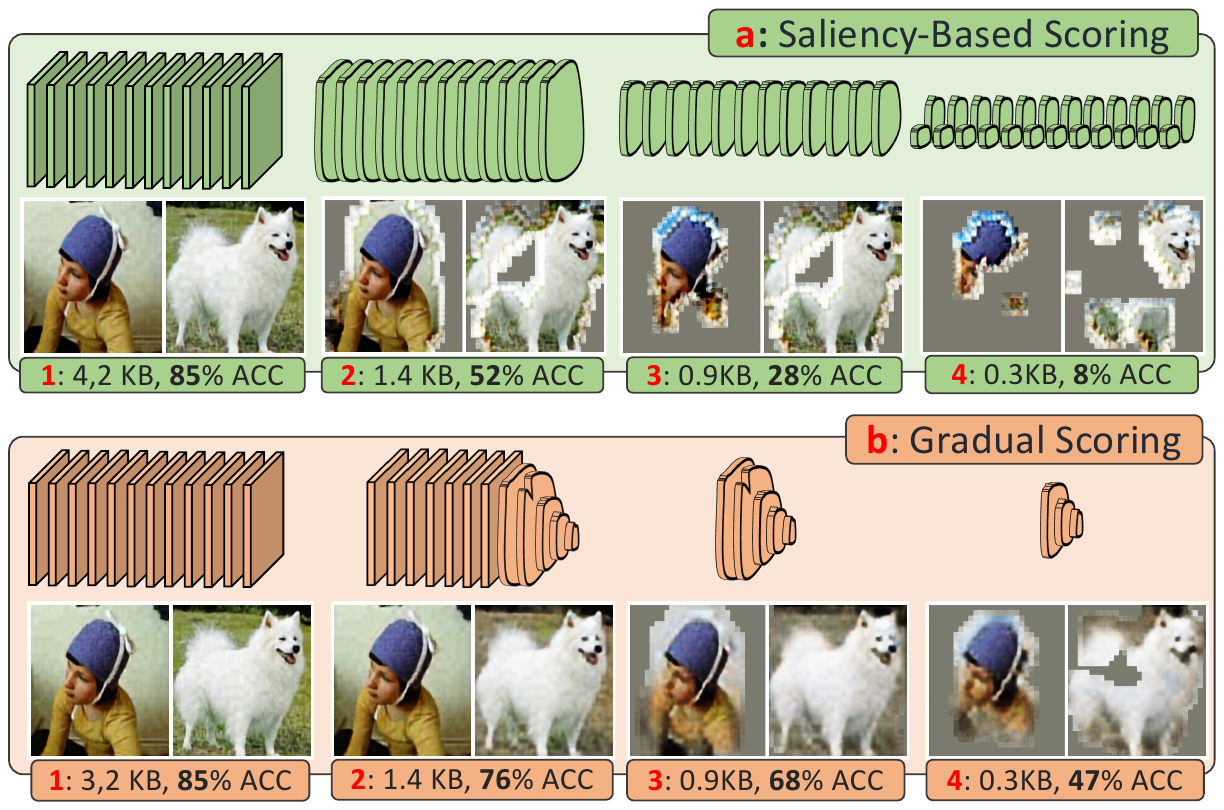}
   \caption{Reconstruction outputs and Top5-Acc (on ImageNet1000) at different data availability levels when we only use the saliency map compared to combining it with Gradual Scoring.}
   \label{fig:dropping}
   % \vspace{-2em}

\end{figure}

\begin{figure}[tb]
    \centering 
    \includegraphics[width=0.5\textwidth]{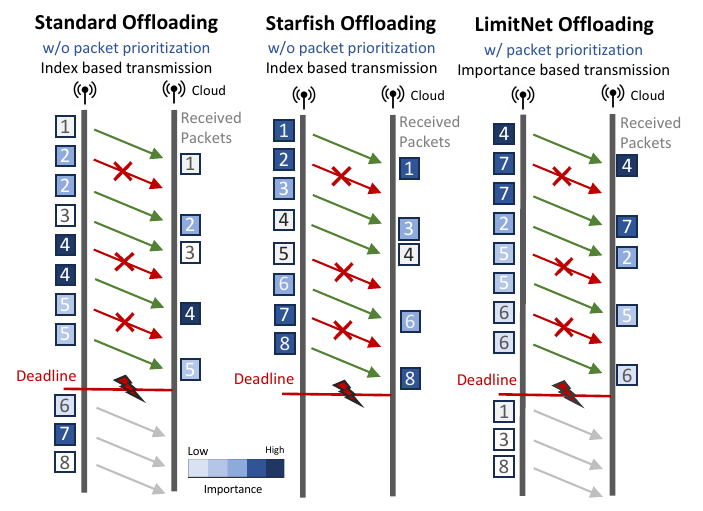}
    \caption{\note{A high-level illustration of \name's transmission-retransmission policies. In conventional offloading schemes, we transmit the data order of its index, and in the case of packet loss, either we skip (like Starfish) or retransmit data with the same policy (standard offloading). \name, in contrast, transmits and retransmit packets based on their importance. Thereby, it ensures that the most important packets arrive first, essential for graceful progressive decoding.}}
    \label{fig:transmission}
    % \vspace{-1.5em}
\end{figure}

\subsection{Saliency Detection} \label{Saliency Detection}
\label{Design:Intra-Channel}
After data encoding, we need to detect the important parts of the input image to add priority to their corresponding encoded data. 

There are several ways to detect the important parts, such as ROI detection, Explainable-AI, and Saliency Detection, see Section~\ref{Related Work}. However, running these models on embedded devices is often infeasible due to their high computational and resource costs, comparable to or even greater than running a full classifier. Despite the hardware limitations, we leverage the benefits of edge computing by employing a lightweight model that \textit{mimics} the output of a complex saliency detector. 
We design a lightweight, 4-layer network $\text{{SalDet}}_{\theta_{SalDet}}$ for saliency detection and train it via \textit{Knowledge Distillation (KD)} \cite{hinton2015distilling} using BASNet \cite{qin2021boundary}, one of the state-of-the-art saliency detection models, as teacher model. This branch takes latent $Z^{L \times K \times K}$ and extracts the saliency map $I^{K \times K}$, see Eq.~\ref{eq:2}. 

\begin{equation}
Z^{L \times K \times K} \rightarrow \text{SalDet}_{\theta_{\text{SalDet}}}(Z^{L \times K \times K}) \rightarrow I^{K \times K}
  \label{eq:2}
\end{equation}

Our saliency detection branch, ($\theta_{\text{SalDet}}$), has only 5k parameters, which is 0.001\% of the parameters of the original saliency model.
Although our lightweight model does not achieve the same level of performance as BASNet, the resulting importance map provides essential information about the important regions, which greatly enhances the model's accuracy, as our evaluation in Section~\ref{Evaluating Detail} shows.

\begin{table*}[tb]
  \caption{Training phases of \name}
\vspace{-1em}
  \label{tab:traning}
  \begin{tabular}{ccccc}
    \toprule
    \text{Phase} & \text{Epochs} & \text{LR} & \text{Loss Function} & \text{Traning Notes}\\
    \midrule
    1 & 100 & 0.001 & Rec Loss + Saliency Loss & \text{KD for the saliency detection, training with Gradual Scoring }\\
    2 & 6 & 0.00005 & CLS Loss & \text{Training with Gradual Scoring, freezing the CLS}\\
    \bottomrule
  \end{tabular}
\end{table*}

\subsection{Gradual Scoring: The Devil is NOT in the Details!} \label{Gradual Scoring}
\note{After extracting the saliency map, a na\"ive solution is to use this map to score each encoded data point. However, relying only on the saliency map leads to fragmented and blocky outputs and hence, poor accuracy. For instance, examples $a2$, $a3$, and $a4$ in Fig.~\ref{fig:dropping} illustrate the decoder outputs with varying levels of available data. As shown in these examples, fragmentation occurs since we only transfer the parts with the highest score in the saliency map, focusing only on the foreground and sending the same locations across the latent for all filters.
Consequently, the decoder can reconstruct these parts only, which hinders the classification model's ability to accurately identify objects due to the \textit{absence of contextual information} \cite{cai2019end}.
To effectively use saliency, we introduce a Gradual Scoring algorithm, which enables the model to \textit{learn and choose} how much background, i.e., context, it needs during the training for better classification results.
Our results show that incorporating Gradual Scoring leads to significantly improved classification accuracy. For instance, examples $a3$ and $b3$ in Fig.~\ref{fig:dropping} show that for the same data size, gradual scoring leads to 40 p.p. higher accuracy on the ImageNet1000 \cite{deng2009imagenet}.}
\\

We design the Gradual Scoring mechanism inspired by the gradual ordering in TailDrop \cite{koike2020stochastic}, which uses dropout to order the information in the latent.
The Gradual Scoring function takes the saliency map $I^{K \times K}$ and by adding a \textit{descending constant value}, $G_{Factor}$, produces a $S^{L \times K \times K}$ tensor, which shows the importance score of each data point in the latent; see Eq.~\ref{3}.

\begin{equation} \label{3}
I^{K \times K} \rightarrow \text{GS}(I^{K \times K}) \rightarrow S^{L \times K \times K} 
\end{equation}

\noindent
As illustrated in Fig.~\ref{fig:scoring}, the score of each latent data point is calculated as follows:

\begin{equation}
S_{i, [0:K], [0:K]} = I_{[0:K], [0:K]} + G_{Factor} \times i, \quad \forall i \in \{0, 1, 2, \ldots, L\}\\
  \label{eq:3}
 \end{equation}
 
To get the best results, we need to employ Gradual Scoring in the training pipeline. To do so, in each training step, we randomly select a value $p = \mathcal{U}(0, 100)$, and zero out the $p \%$  of the latent $Z^{L \times K \times K}$ with the \textit{lowest score} in the score tensor $S^{L \times K \times K}$; see Eq.~\ref{5} and Eq.~\ref{6}.

\begin{equation} \label{5}
Z^{L \times K \times K}, S^{L \times K \times K} \xrightarrow{\text{Dropping}} Z'^{L \times K \times K}
\end{equation}

\begin{equation} \label{6}
{Z'}_{i,j,k} = \begin{cases}
{Z}_{i,j,k} & \text{if } S_{i,j,k} \geq p^{th}\text{largest value of } S^{L \times K \times K}\\
0 & \text{otherwise}
\end{cases}
\end{equation}

\noindent
$Z'^{L \times K \times K}$ represents the modified version of $Z^{L \times K \times K}$ that contains the $p \%$ of the highest important scores.

\subsection{Offloading, Decoder and Classifier} \label{Decoder, and Classifier}
After encoding, we quantize the latent data with 6 bits. We also downsize the saliency map to an $8 \times 8$ representation and then quantize it with 5 bits\note{, as we do not lose a lot of accuracy and add considerable compression.} 
\review{We initiate the offloading process by first transmitting the saliency map before sending the encoded data. The overhead of transmitting this map is a maximum of 40 bytes, which is practically negligible. For instance, in an LPWAN network with a throughput of 5 KB/s, it takes less than 1 ms to transmit. This step is crucial as the decoder requires precise placement information for reconstructing each data point. }
On the cloud side, after receiving the compressed image data, we fill any unreceived latent values with zero based on the saliency map and reconstruct the latent $\hat{Z}^{L \times K \times K}$.
While we prioritize minimizing computational costs during encoding, we are less concerned about the costs of decoding. We employ a decoder ${DEC}_{\theta_{DEC}}$ comprising a 5-layer convolution with different kernel sizes (7, 5, and 3). These varying kernel sizes allow the decoder to capture features at different scales and reconstruct the finer details of the original image, see Fig~\ref{fig:arch}. This decoder is responsible for reconstructing the original image ${X}^{C \times H \times W}$. Once the image is reconstructed, we feed it to the vision model ${DEC}_{\theta_{DEC}}$ and predict the label $\hat{y}$; see Eq.~\ref{7} and Eq.~\ref{8}. Specifically, in this paper, we utilize EfficientNet-B0 \cite{tan2019efficientnet} and YOLOv5 \cite{redmon2016yolo}, depending on the evaluation scenario. 

\begin{equation} \label{7}
{Z'}^{L \times K \times K} \xrightarrow{\text{Quant., Huffman Enc.}} \hat{Z'}^{L \times K \times K}
\end{equation}
\begin{equation} \label{8}
\hat{Z'}^{L \times K \times K} \xrightarrow{\text{Lat. Rec.}} \hat{Z}^{L \times K \times K} \xrightarrow{{DEC}_{\theta_{DEC}}} 
\hat{X}^{C \times H \times W} \xrightarrow{{CLS}_{\theta_{CLS}}} \hat{y}
\end{equation}

\noindent
\subsubsection{Prioritized Packet Transmission-Retransmission} ~\label{packet loss}
In \name, we offload packets based on their importance score produced by Gradual Scoring. However, as discussed in Section~\ref{LPWAN}, LPWAN links are highly dynamic and often have a high packet loss rate, necessitating retransmissions.

Due to its content-aware progressive bitstream, the importance of each packet in \name is known.
Therefore, we transmit and -- if needed -- retransmit packets in order of importance until an application-specific deadline is reached or our duty cycle budget is over. 
As a result, this ensures that the receiver always receives the most important data in the case that we cannot transmit the entire latent. 
Looking at Fig.~\ref{fig:transmission}, through \name's prioritized offloading, critical packets (4, 7, 2, 5, and 6) are delivered to the cloud before the deadline. In contrast, standard offloading schemes send packets based on their index order without considering their importance, resulting in the cloud missing crucial packets (e.g., packets 6 and 7).
Although Starfish has regional importance for mitigating packet loss, it also lacks packet prioritization capabilities. Thus, similar to the conventional offloading models, it transmits packets in random order and potentially loses crucial packets (e.g., packets 2 and 7).

\review{
\begin{figure}[tb]
  \centering
   \includegraphics[trim=45 0 0 0, clip, width=.49\textwidth]{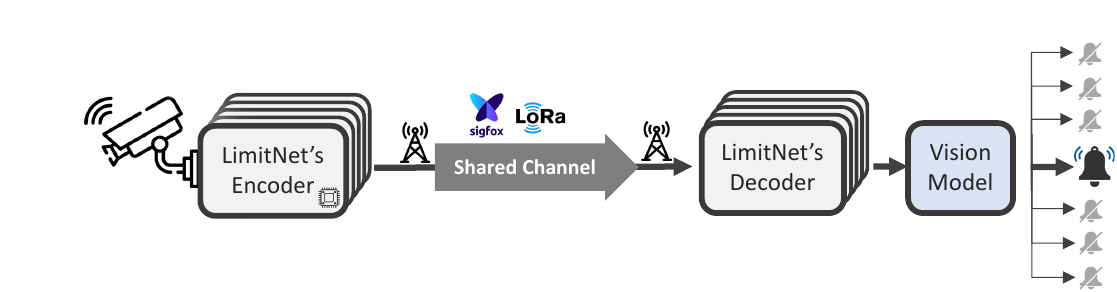}
   \caption{\review{A high-level application scenario for \name. In this scenario, multiple cameras transmit their images through a shared LPWAN link.}}
   \label{fig:APPLICATION}
   % \vspace{-1em}
\end{figure}
}

\review{\subsection{Application Scenario}
Distributed IoT cameras, such as alarm systems, utilize a shared LPWAN connection for transmitting data to the cloud, necessitating limited duty cycles.
In conventional approaches, inference needs to wait until all data is received, where it takes approximately 100 seconds to send a 5KB image through a 5 KB/s LoRa link with a 1\% duty cycle.
However, in \name, each node compresses images progressively, sorts encoded data by importance, and transmits them accordingly within its duty cycle. On the cloud end, after each cycle, the decoder can reconstruct the image, conduct inference for each camera, and activate the alarm system if needed, without needing to wait for the complete data to be received; see Fig.~\ref{fig:APPLICATION}.
}

\section{Evaluation}\label{Evaluation}
In this section, we evaluate the performance of \name and compare it to state-of-the-art image compression and offloading models. We focus on vision tasks like classification and object detection, use accuracy and mAP as evaluation metrics, and assess system-level benchmarks on MCUs as follows:

\noindent
\textbf{Accuracy vs. Data Size:} 
We evaluate \name's accuracy on different compression ratios and data sizes and compare it to SOTA models.

\noindent
\textbf{Saliency Detection Evaluation:} 
We evaluate the saliency branch and the Gradual Scoring algorithm to analyze how well they identify important parts. 

\noindent
\textbf{System-level Benchmarks on MCU:} We implement \name on STM32F7 (Cortex-M7) and nRF5340 (Cortex-M33) and evaluate system-level benchmarks. In our experiments, we focus on Flash usage, RAM usage, execution time, and energy consumption. Furthermore, we evaluate \name and SOTA models' performance under LoRaWAN networks.

\noindent
\note{\textbf{MS-SSIM and PSNR}: While it is common to use visual metrics such as MS-SSIM and PSNR when comparing image compression models, \name only produces a partial reconstruction (see Fig. \ref{fig:demo}). As these metrics measure the average difference of all pixels, the result will be significantly biased to the missing values. However, the main objective of \name is classification accuracy and not the perceptual quality. Therefore, we do not evaluate these metrics.}

\review{Note that, for simplicity, we evaluate the performance of \name under the assumption of a single image in the transmission queue, as is common in the literature \cite{huang2022real, rethinking, hu2020starfish, huang2020clio}. However, \name can also be extended to handle multiple images, either in parallel or sequentially, which we leave for future work.}

\subsection{Experimental Setup}

\subsubsection{Implementation} 
We implement \name in TFLite-Micro \cite{tflite-micro} and Zephyr RTOS \cite{zephyr} for STM32F7 \cite{STM32F7Series} and nRF5340 \cite{nRF5340} MCUs. The STM has 2 MB of Flash, 1 MB of RAM, and a dual-core configuration consisting of an ARM Cortex-M7 and an ARM Cortex-M33, which can be clocked up to 480 MHz and 240 MHz, respectively. 
We deploy \name on the Cortex-M7. 
The nRF has 1 MB Flash, 512 kB RAM, and a dual ARM Cortex-M33 core that can be clocked at 128 and 64 MHz.
Both MCUs provide accelerated integer inference via DSPs and CMSIS-NN. To effectively utilize \name and ensure that the model's weights do not excessively consume memory resources, we apply post-training integer quantization to \name's image encoder, i.e., we quantize the weights to 8-bit integer representations. 
\name is available as open source\footnote{\review{https://github.com/ds-kiel/LimitNet}}.

\noindent
\note{\subsubsection{Datasets} We evaluate \name on three public datasets: To show the maximum capacity of our model, we train and evaluate \name on ImageNet1000 \cite{deng2009imagenet}. We also fine-tune and evaluate our model on CIFAR100 \cite{cifar100}. To investigate the performance of \name on harder tasks such as object detection, we also evaluate the pre-trained ImageNet1000 model on the COCO dataset \cite{cocodataset} by using YOLOv5 \cite{redmon2016yolo}. For all datasets, we resize the input images to $224 \times 224$.}

\subsubsection{Training and hyperparameters} Our training process has two phases. At first, we train the autoencoder and the saliency branch jointly using Gradual Scoring with $G_{Factor}=0.2$. 

In the second phase, we stitch the classification model after the decoder and train the whole network on classification loss, see Table~\ref{tab:traning}.

\begin{figure*}
    \centering
    \begin{subfigure}[b]{0.33\textwidth}
        \centering
        \includegraphics[width=1\textwidth]{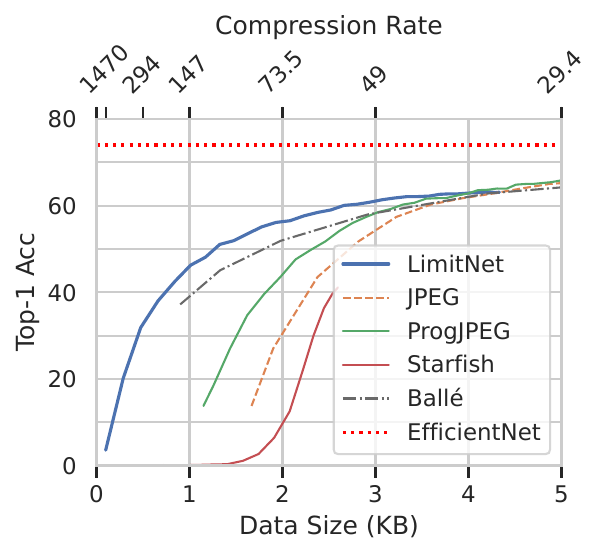}
        \caption[]%
        {{\small Data Size vs. Accuracy on ImageNet}}    
        \label{fig:ACC-ImageNet}
    \end{subfigure}
    \begin{subfigure}[b]{0.33\textwidth}
        \centering 
        \includegraphics[width=1\textwidth]{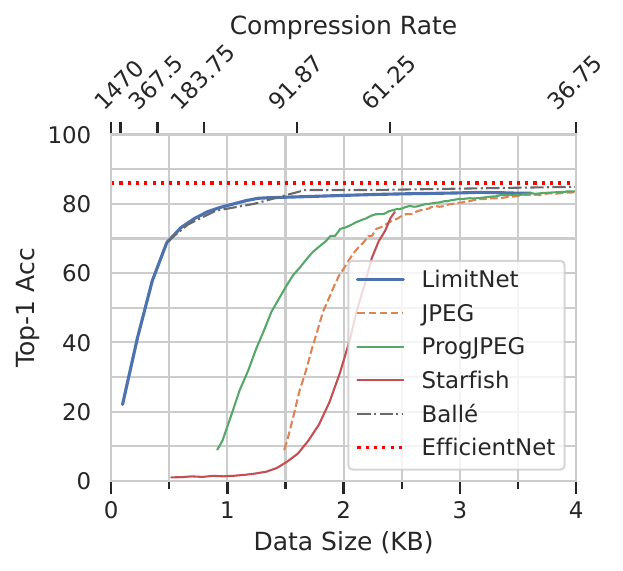}
        \caption[]%
        {{\small Data Size vs. Accuracy on CIFAR}}      
        \label{fig:ACC-CIFAR}
    \end{subfigure}
    \begin{subfigure}[b]{0.33\textwidth}
        \centering
        \includegraphics[width=1\textwidth]{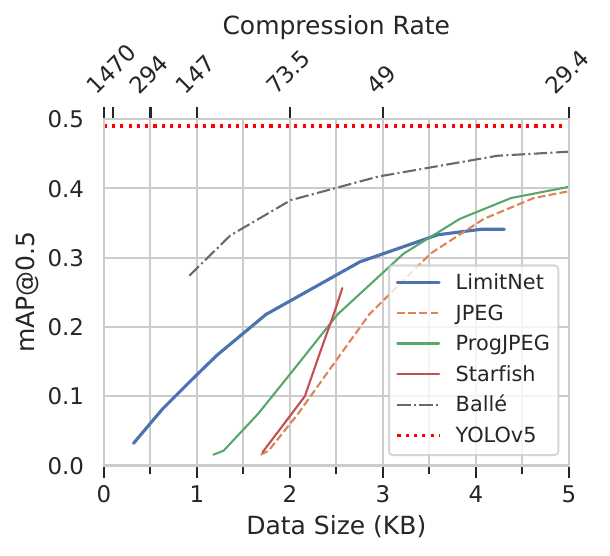}
        \caption[]%
        {{\small Data Size vs. mAP@0.5 on COCO}}    
        \label{fig:mAP-COCO}
    \end{subfigure}
   \caption{Performance evaluation of \name compared to JPEG%\cite{wallace1991jpeg}
   , ProgJPEG %\cite{wallace1991jpeg}
   and Starfish %\cite{hu2020starfish}
   on ImageNet1000%\cite{deng2009imagenet}
   , CIFAR100 %\cite{cifar100} 
   and COCO%\cite{cocodataset}
   . We also compare it to the SOTA image compression model, Ballé \etal%~\cite{balle2018variational}
   , which has 165 times more parameters than \name and is not executable on MCUs.}
   \label{accuracy}
\end{figure*}

\subsection{Compression Efficiency} \label{Compression Efficiency}

\noindent
\textbf{Baselines:} In this section, we compare the compression efficiency of \name to three sets of baselines:
(1) resource-efficient compression models deployable on embedded MCUs: Starfish \cite{hu2020starfish} as the only existing image compression model that can handle partial data availability, JPEG \cite{wallace1992jpeg} as the standard baseline, and the progressive version of JPEG (referred as ProgJPEG) as the progressive standard baseline.
(2) SOTA image compression models: Ballé \etal \cite{balle2018variational} as the non-progressive image compression which has 165 times more parameters and 50 times more GFLOPs compared to \name, see Table \ref{tab:params_flop}. \review{Although there are newer and improved models, we choose Ballé as it is easy to evaluate, well-documented, and has comparable performance to SOTA.}
(3) Non-progressive offloading models: BottleNet++ \cite{shao2020bottlenet++} as the standard benchmark and DeepCOD \cite{yao2020deep} as SOTA.

We found that post-training integer weight quantization has less than 0.01 p.p. effect on the results, which is negligible. Given this, we only present the non-quantized results.

\begin{table}
\caption{Comparing GFLOPs and the \#parameters of \name with Ballé \etal%\cite{balle2018variational}
, the state-of-the-art image compression model, and Starfish%\cite{hu2020starfish}
, a MCU-based deep image compression model.}
\vspace{-1em}
\begin{tabular}{|l|l|l|l|l}
\hline
\textbf{Model} & \textbf{\#GFLOPs} & \textbf{\#Params} \\
\hline
\textbf{\name} & \textbf{0.004} & \textbf{15K} \\
% DeepCOD & 0.0003 & $96$  \\
Starfish & 0.77 & $78K$  \\
Ballé \etal & $1.95$ & $2.5M$ \\

\hline
\end{tabular}
\label{tab:params_flop}
\vspace{-1em}

\end{table}

\noindent
\textbf{Metrics:} In our evaluation, we use these metrics for comparison:
\subsubsection{Data size vs. Accuracy/mAP} We evaluate \name and other benchmarks by feeding their decoded image into EfficientNet-B0 \cite{li2020efficient} as a backend model for classification. 
Fig.~\ref{fig:ACC-ImageNet} and Fig.~\ref{fig:ACC-CIFAR} plot Top-1 classification accuracy and show that \name consistently outperforms the baseline models. For example, on CIFAR100, with 1 KB of data, \name achieves 80\% accuracy while ProgJPEG achieves 16\% accuracy. 
%, which is very useful when dealing with extremely limited and dynamic bandwidths.
At high compression rates, \name even outperforms Ballé \etal on ImageNet1000 and achieves an on-par performance on CIFAR100. We also evaluate \name on object detection using YOLOv5 \cite{redmon2016yolo} as the backend model on the COCO dataset. As shown in Fig.~\ref{fig:mAP-COCO}, \name outperforms the baselines at high compression rates when compressed image sizes are 3.5 KB and smaller.
\note{This superior performance primarily stems from \name's ability to reconstruct the image components crucial for classification. In contrast, other progressive models like Starfish and ProgJPEG are content-agnostic and transmit images without prioritizing the crucial regions regarding classification.}

\begin{figure}[tb]
  \centering
        \includegraphics[width=.47\textwidth]{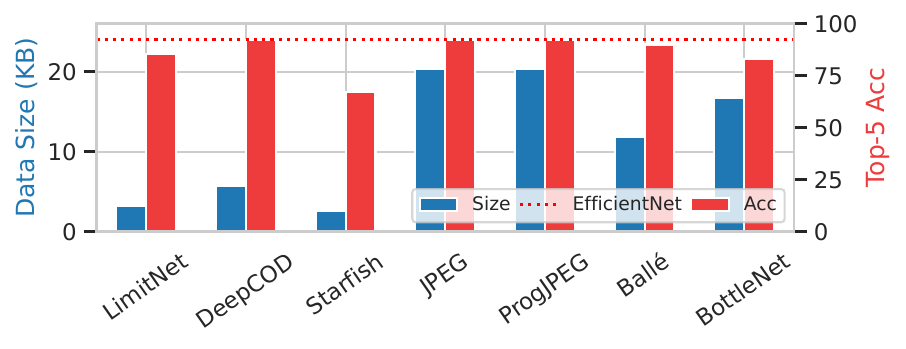}
    \vspace{-1em}
   \caption{\name performance evaluation, compared to non-progressive offloading models (DeepCOD \cite{yao2020deep}, BottleNet++ \cite{shao2020bottlenet++}, and Ballé \etal \cite{balle2018variational}) and other benchmarks \cite{hu2020starfish, wallace1991jpeg}, when 100\% of the encoded data is available. \name achieves comparable performance to these non-progressive models despite being a progressive encoder.}
   \label{fig:MaxPerormance-ImageNet}
   \vspace{-1.5em}

\end{figure}

Additionally, we evaluate \name by comparing it with non-progressive offloading models such as DeepCOD \cite{yao2020deep} and BottleNet++ \cite{shao2020bottlenet++} when 100\% of the data is available for decoding. We choose these models' first offloading point, ensuring implementation feasibility on MCUs. As shown in Fig.~\ref{fig:MaxPerormance-ImageNet}, \name achieves comparable performance to these non-progressive models, despite being a progressive encoder\footnote{As common, when comparing to DeepCOD, we plot Top5-Acc.}.

\subsubsection{BD-Rate/BD-Acc/BD-mAP} 
To summarize the results on compression efficiency, we utilize the Bjontegaard Delta (BD) \cite{bjotegaard2001calculation}, a metric for comparing various model performances over different data sizes:

\begin{itemize}
    \item  \textbf{BD-Rate} shows, on average, how much \textit{bandwidth} a baseline saves over another baseline for a given \textit{quality} in percent.
    \item \textbf{BD-Acc} shows, on average, how much a baseline improves the \textit{accuracy} in percentage points (p.p.) over another baseline for a given \textit{data size}.
    \item \textbf{BD-mAP} shows, on average, how much \textit{mAP} a baseline improves over another baseline for a given \textit{data size}.
\end{itemize}

\begin{table*}[ht]
\centering
\caption{Comparison of \name with baselines in terms of BD-Rate (in percent), BD-Acc (in percentage points), and BD-mAP@50. Each number in each column shows the improvement of \name compared to the corresponding model.}
\vspace{-1em}
\begin{tabular}{|c|c|c|c|c|c|c|}
\hline
\multirow{2}{*}{\textbf{Model}} & \multicolumn{2}{c|}{\textbf{ImageNet1000}} & \multicolumn{2}{c|}{\textbf{CIFAR100}} & \multicolumn{2}{c|}{\textbf{COCO}} \\ \cline{2-7}
 & BD-Rate (\%) & BD-Acc (p.p.)  & BD-Rate (\%)  & BD-Acc (p.p.)  & BD-Rate (\%)  & BD-mAP (mAP@50)  \\ \hline
\textbf{ProgJPEG} & 61.24 & 14.01 & 83.68 & 18.01 & 42.45 & 0.1 \\ \hline
\textbf{JPEG} & 69.02 & 15.41 & 87.7 & 15.21 & 53.64 & 0.12 \\ \hline
\textbf{Starfish} & 85.09 & 60.33 & 89.81 & 72.26 & 55.02 & 0.15 \\ \hline
\end{tabular}
\label{tab:BD}
\end{table*}

Table \ref{tab:BD} presents the BD metrics for \name compared to JPEG%\cite{wallace1991jpeg}
, ProgJPEG, and Starfish. %\cite{hu2020starfish}.
\textit{Accuracy/mAP results}: for a given data size, \name on average achieves up to 14.01 percentage points higher accuracy compared to the SOTA on the ImageNet1000 dataset, 18.01 percentage points accuracy improvement on the CIFAR100 dataset, and 0.1 mAP@0.5 improvement on the COCO dataset.
(2) \textit{Rate results}: For a given accuracy or mAP@50, \name on average saves 61.24\% bandwidth compared to the SOTA on ImageNet1000 dataset, 83.68\% on the CIFAR100 dataset, and 42.25\% on the COCO dataset.

\subsection{Evaluating \name in Detail} \label{Evaluating Detail}
In this section, we conduct a comprehensive evaluation of each of \name's components to measure the benefits of content-aware encoding, assess the impact of our Gradual Scoring mechanism using different $G_{Factor}$ values, and evaluate the performance of our saliency detection branch.

\subsubsection{Advantages of Content-Aware Encoding} 
\begin{figure}
    \centering
   \begin{subfigure}[t]{0.235\textwidth}
        \centering 
        \includegraphics[trim=5 5 5 5, width=1.02\textwidth]{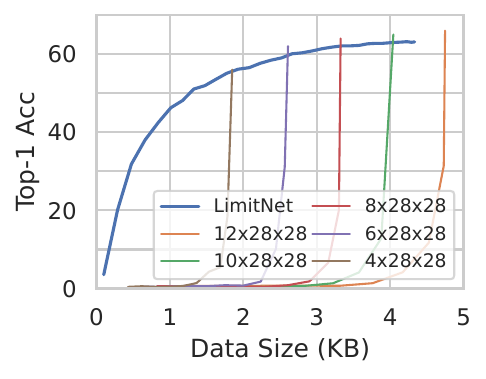}
        \caption[]%
        {{\small}}    
        \label{fig:DropAE}
    \end{subfigure}
    \begin{subfigure}[t]{0.235\textwidth}
        \centering
        \includegraphics[trim=5 5 5 5, width=1.02\textwidth]{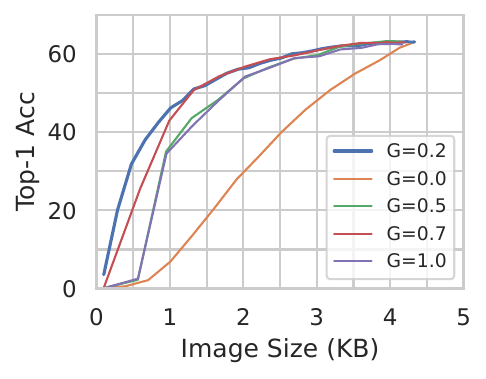}
        \caption[]%
        {{\small}}    
        \label{fig:ScorePlot}
    \end{subfigure}
    \vspace{-1em}
   \caption{(a) Comparing \name and Fixed-Rate AEs results on incomplete data: \name outperforms Fixed-Rate AEs when we do not have all the encoded data for inference. (b) Evaluating different $G_{Factor}$. $G_{Factor}=0.2$ achieves the best performance, effectively balancing object details and context in the reconstructed image.}
   % \vspace{-1.5em}
\end{figure}

We begin by assessing the impact of progressive content-aware encoding on the model results. 
For this, we train a set of non-progressive autoencoders with different latent sizes and run the inference on incomplete data. Fig.~\ref{fig:DropAE} illustrates that a normal, i.e., non-progressive, autoencoder shows a significant drop in accuracy when reconstructing the image with incomplete data, while in \name, the accuracy gracefully degrades due to its progressive nature and having a content-aware encoding.

\subsubsection{Gradual Scoring}

\note{In this section, we evaluate the effect of Gradual Scoring on classification accuracy. 
We train our model with different $G_{Factor}$ and measure its accuracy on the ImageNet1000 dataset. 
As Fig.~\ref{fig:ScorePlot} shows, if we set $G_{Factor}$ = 0, i.e., using the saliency map naively, it leads to a poor accuracy since it produces a blocky output. On the other hand, if we set $G_{Factor}$ = 1.0, the model diminishes the effect of the saliency map, which hinders it from achieving its maximum accuracy. By choosing an intermediate value for the $G_{Factor}$, we allow the model to balance foreground and background in the reconstructed image. Our experiments show that $G_{Factor}$ = 0.2 produces the best results; see Fig.~\ref{fig:ScorePlot}.}

\begin{figure*}[tb]
  \centering
   \includegraphics[trim=30 130 30 130, clip, width=1.0\textwidth]{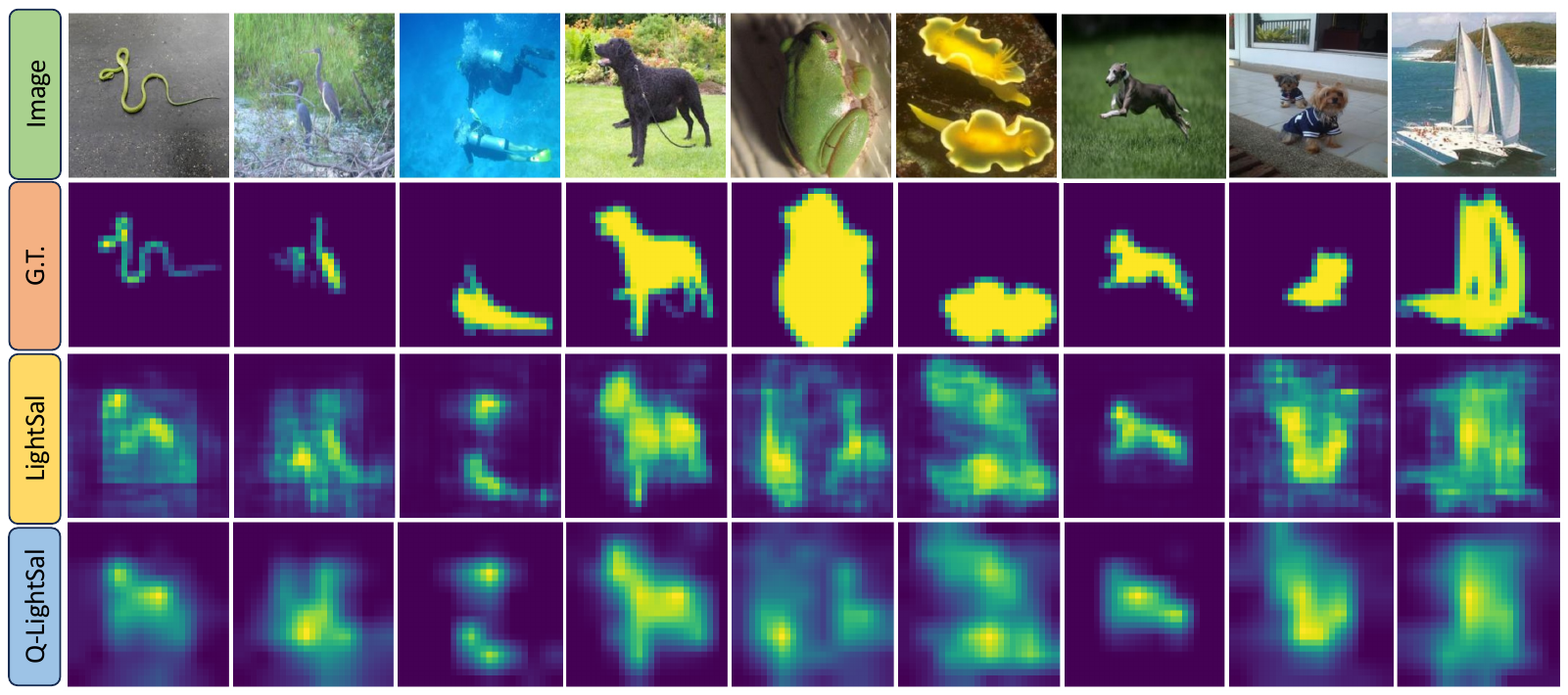}
   \caption{Comparing the input image, saliency ground truth (G.~T.), \name's saliency output with and without quantization, denoted as Q-LightSal and LightSal, respectively. While the saliency detection branch does not achieve the same level of preciseness as the ground truth, it detects the parts of the image that include the target object, which bumps up the accuracy when we prioritize these sections in progressive offloading.}
   \label{fig:sal_output}
\end{figure*}

\subsubsection{Evaluation of Saliency Detection} \label{Evaluation of Saliency Detection} 
As discussed in Section~\ref{Decoder, and Classifier}, in the quantization step, we downsize the $32 \times 32$ maps to $8 \times 8$ for better compression and encode these in 5 bits. 
Therefore, when evaluating the saliency maps, we also need to evaluate the impact of this quantization.
In Fig.~\ref{fig:sal_output}, we compare the input image, saliency ground truth, and \name's saliency output with and without quantization. 
We use BASNet \cite{qin2021boundary}, i.e., our teacher model, as ground truth. 
As this figure shows, the saliency detection branch is not as accurate as the ground truth; nevertheless, it can identify the \textit{area} of the important parts of the image. For example, in the fourth column, although the saliency can not precisely localize the dog, it detects the parts of the image that include the target object, which bumps up the accuracy when we prioritize these sections in progressive offloading.

\begin{table*}
    \caption{Resource requirements and inference time of both \name's encoder and the saliency detector as float32 and  quantized int8 models with and without DSP acceleration on nRF5340 and STM32F7 MCUs.}
    \begin{tabular}{|c|c|c|c|c|c|c|c|c|c|}
    
    \hline
    \multicolumn{2}{|c|}{} & \multicolumn{4}{c|}{\textbf{nRF5340 MCU (Cortex-M33)}} & \multicolumn{4}{c|}{\textbf{STM32F7 MCU (Cortex-M7)}} \\
    \cline{3-10}
    \multicolumn{2}{|c|}{} & \multicolumn{2}{c|}{\textbf{without DSP accel.}} & \multicolumn{2}{c|}{\textbf{with DSP accel.}}  & \multicolumn{2}{c|}{\textbf{without DSP accel.}} & \multicolumn{2}{c|}{\textbf{with DSP accel.}} \\
    \cline{3-10}
    \multicolumn{2}{|c|}{} & int8 & float32 & int8 & float32 & int8 & float32 & int8 & float32 \\
    \hline
    \multirow{3}{*}{\rotatebox[origin=c]{90}{\textbf{Enc.}}} & exe time (ms) & 45,590 & - & \textbf{\textbf{1,969}} & - & 4,882 & - & \textbf{\textbf{237}} & - \\
    % \cline{2-10}
    & RAM (KB) & 344 ($33\%$) & - & \textbf{344 ($\textbf{33\%}$}) & - & \textbf{350 ($\textbf{17\%}$)} & - & 360 ($17\%$) & - \\
    % \cline{2-10}
    & Flash (KB) & \textbf{94 ($\textbf{9\%}$}) & - & 100 (10\%) & - &  \textbf{91 ($\textbf{4\%}$)} & - & 107 (5\%) & - \\
    \hline
    \multirow{3}{*}{\rotatebox[origin=c]{90}{\textbf{Sal.}}} & exe time (ms) & 4,972 & 1,046 & \textbf{\textbf{220}} & 981 & 522 &  576  & \textbf{\textbf{23}} & 655 \\
    % \cline{2-10}
    & RAM (KB) & 26 ($2\%$) & 99 ($10\%$) & \textbf{26 ($\textbf{2\%}$)} & 99 ($10\%$) & \textbf{24 ($\textbf{1\%}$)}  & 99 ($5\%$) &  26 ($1\%$) &  99 ($5\%$) \\
    % \cline{2-10}
    & Flash (KB) & \textbf{96 ($\textbf{9\%}$)} & 111 ($11\%$)  & 102 ($10\%$) & 117 ($11\%$) & \textbf{93 ($\textbf{4\%}$)} & 109 ($5\%$) & 107 ($5\%$) & 122 ($6\%$) \\
    \hline
    \end{tabular}
    
    \label{tab:nrf}
\end{table*}

\subsection{System-Level Benchmarks} \label{System-Level Benchmarks}

Next, we evaluate the system-level performance of \name, including current consumption, energy tracing, resource consumption, and performance under limited network bandwidths such as LoRaWAN. \review{Although Starfish is tailored for limited hardware settings, it cannot run on the selected MCUs due to its extensive RAM usage. Therefore in this section, we only evaluate \name and JPEG.}

\subsubsection{Resource Consumption} 
In Table~\ref{tab:nrf}, we report the resource requirements and inference time for the float32 model and the quantized int8 model with and without DSP support for NN inference acceleration on nRF5340 and STM32F7 MCUs.
Only the int8 version of the encoder is deployable on MCU due to space limitations, while we can deploy the saliency detection branch with either float32 or int8 model weights. 
For int8, the encoder consumes 344 KB RAM (33\% of RAM), about 100 KB Flash (10\% of Flash), and takes 1,969 ms to run on nRF with DSP acceleration enabled and 360 KB RAM (17\% of RAM), 107 KB Flash (5\% of Flash), and 237 ms to run on STM with DSP acceleration enabled. 
Also, the saliency detection branch as quantized int8 model needs 26 KB RAM (2\% of RAM), 102 KB Flash (10\% of Flash), and 220 ms to run on nRF5340 with DSP acceleration enabled, and 26 KB RAM (1\% of RAM), 107 KB Flash (5\% of Flash), and 23 ms to run on STM with DSP acceleration enabled.
Overall, \name compresses images within 260 ms on the faster Cortex-M7 while it needs about 2,189 ms on the slower Cortex-M33.  

\subsubsection{Current Consumption and Energy Trace}
\begin{figure}[tb]
    \centering 
    \includegraphics[trim=10 25 10 10, width=0.47\textwidth]{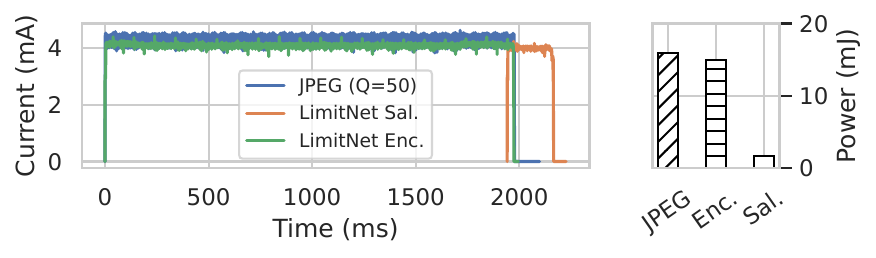}
    \caption{Energy and current consumption of \name's encoder and saliency detection compared to JPEG (for one predefined constant quality=50) on nRF5340 with $V=1860 mV$.}
    \label{fig:Power}
    % \vspace{-1em}
\end{figure}

To gain insights into the energy demands of \name, we measure energy consumption on the nRF5340 MCU and compare it to JPEG ($Q=50$).
Fig.~\ref{fig:Power} shows that the encoding takes about $11\%$ longer, i.e., about $220$ ms longer, for \name when compared to JPEG on the nRF. 
Also, on the STM, \name encoding takes $260$ ms while JPEG needs $249$ ms, i.e., \name requires about $4\%$ more time than JPEG.
Regarding energy consumption, \name consumes $16.6$ mJ, which is 0.65 mJ or $4\%$ higher than JPEG ($Q=50$) on nRF, see  Fig.~\ref{fig:Power}.

\note{For this experiment, we use an implementation of JPEG that has been optimized for Cortex-M33 \cite{github:JPEGEncoder4Cortex-M}; however, this implementation does not support progressive encoding while \name does. 
To provide a progressive bitstream, ProgJPEG partitions the DCT coefficient table into multiple segments and encodes each separately. This entails quantization and the creation of Huffman tables, increasing the run-time of ProgJPEG over JPEG. Therefore, compared to \name, we expect that ProgJPEG takes much more time\footnote{\note{While some MCUs, e.g., STM, have a hardware-accelerated JPEG peripheral, others, e.g., nrf5340, do not have one. For a fair comparison, we opted to use an implementation that does not utilize such accelerators.}}.}
 
\subsubsection{Limited Networking Scenarios}

\begin{figure*}
    \centering
    \begin{subfigure}[b]{0.24\textwidth}
        \centering 
        \includegraphics[trim= 6 6 6 6 ,width=1\textwidth]{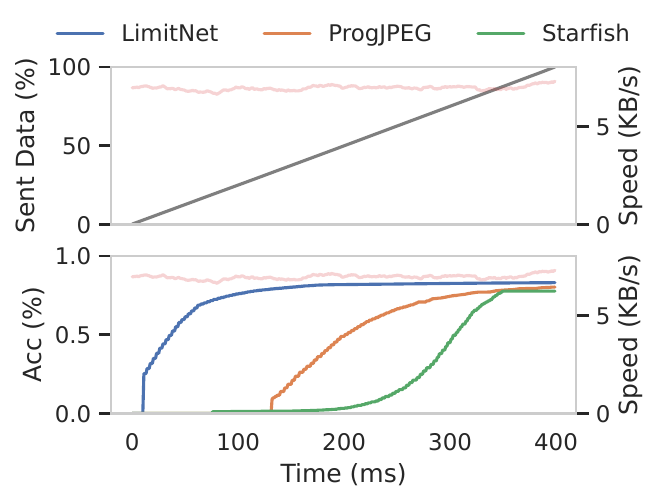}
        \caption[]%
        {{\small 6.25 KB/s - CIFAR100}}    
        \label{fig:Network12_CIFAR}
    \end{subfigure}
    \hfill
    \begin{subfigure}[b]{0.24\textwidth}
        \centering 
        \includegraphics[trim= 6 6 6 6 ,width=1\textwidth]{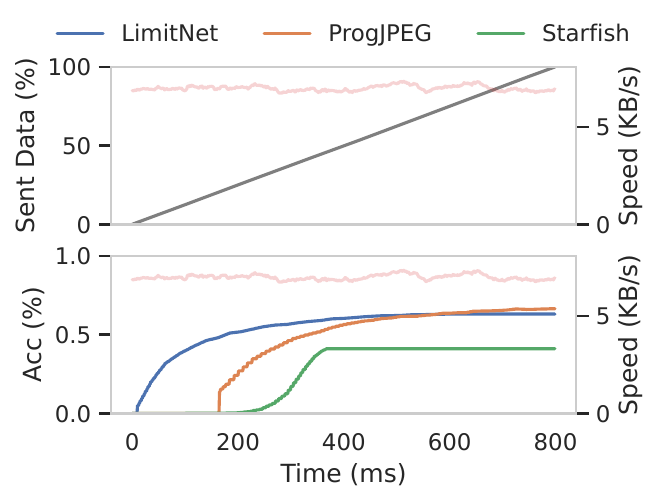}
        \caption[]%
        {{\small 6.25 KB/S - ImageNet1000}}    
        \label{fig:Network12_IMAGENET}
    \end{subfigure}
    \hfill
    \hfill
    \begin{subfigure}[b]{0.24\textwidth}
        \centering 
        \includegraphics[trim= 6 6 6 6 ,width=1\textwidth]{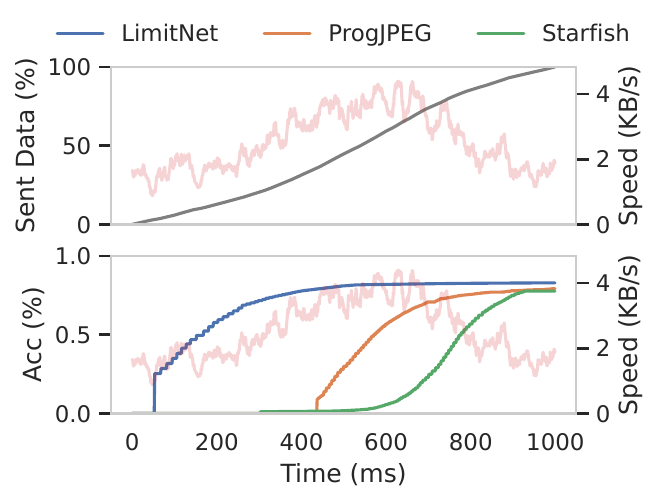}
        \caption[]%
        {{\small Dynamic - CIFAR100}}    
        \label{fig:Network21_CIFAR}
    \end{subfigure}
    \hfill
    \begin{subfigure}[b]{0.24\textwidth}
        \centering 
        \includegraphics[trim= 6 6 6 6,width=1\textwidth]{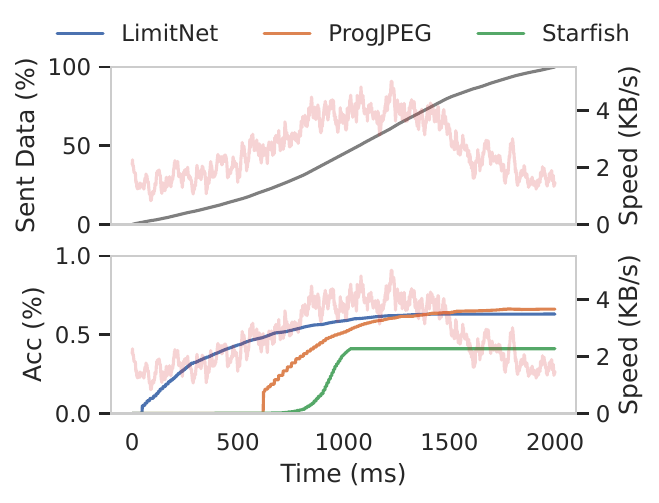}
        \caption[]%
        {{\small Dynamic - ImageNet1000}}    
        \label{fig:Network21_IMAGENET}
    \end{subfigure}
    \hfill
   \caption{\note{\name performance simulation under LoRaWAN \cite{bor2016lora} on ImageNet1000 \cite{deng2009imagenet} and CIFAR100 \cite{cifar100} datasets  compared to Starfish \cite{hu2020starfish} and ProgJPEG \cite{wallace1991jpeg}. We evaluate the performances on each dataset under two different network settings: (a, b) with LoRaWAN's maximum bandwidth and (c, d) with a dynamic bandwidth. These plots show the average outcomes obtained from multiple experiments.}}
\label{fig:Network}
\end{figure*}

To assess the advantages of $\text{\name}$ and compare it to other models in real-world network scenarios, we conduct evaluations for LoRaWAN.
In these experiments, we simulate a network with a gamma distribution under the bandwidth range of LoRaWAN \cite{chaudhari2020lpwan}, see Fig.~\ref{fig:Network}.
In each scenario, we analyze $\text{\name}'s$ performance compared to Starfish \cite{hu2020starfish} and ProgJPEG \cite{wallace1991jpeg}.
We select these two baselines as they are, to the best of our knowledge, the only image compression methods supporting progressive inference while being implementable on MCUs. Our results show that \name exhibits a consistent progressive improvement in accuracy over time and outperforms baselines by a considerable margin; see Fig.~\ref{fig:Network}.
\note{The main factor for this superior performance comes from \name's ability to start reconstruction immediately after sending at most 40 bytes for the saliency map, whereas ProgJPEG needs to send at least the first scan, and Starfish needs to transmit a noticeable amount of data to reconstruct a meaningful image.}

\subsubsection{Impact of packet loss}

\note{As discussed in Section~\ref{LPWAN}, LPWANs often have high packet loss rates, necessitating retransmission of dropped packets. As explained in Section~\ref{packet loss}, in the case of packet loss, \name retransmits packets based on their importance, ensuring that the decoder receives the most important data first. ProgJPEG also sends data in multiple importance-based DCT scans. In Starfish, despite having regional importance distribution for packet loss resiliency, it has no packet prioritization and sends the data in a random order.
In Table~\ref{tab:packet_loss}, we report the impact of different packet loss rates on \name, ProgJPEG \cite{wallace1991jpeg}, and Starfish \cite{hu2020starfish} during a single communication interval. Under different packet loss rates, although \name, Starfish, and ProgJPEG have all sent an equal amount of data, \name outperforms the other models by a significant margin due to its content-aware offloading regime.}

\section{Related Work} \label{Related Work}
The design of \name draws inspiration from a wealth of prior research, including image compression, progressive image compression, compressive sensing, edge offloading, and importance detection methods:

\noindent
\textbf{Image compression: } We categorize image compression models into two groups: classical approaches such as JPEG \cite{wallace1991jpeg}, JPEG2000 \cite{skodras2001jpeg}, JPEG-XR \cite{jpegxr}, WEBP \cite{webp}, BPG \cite{bpg} and recent deep learning-based methods, including Hyperprior \cite{balle2018variational}, Full-Resolution \cite{toderici2017full}, and others \cite{minnen2018joint, cheng2020learned, su2020scalable, Balle2016end}. 
While providing excellent performance, deep models and most of the classical compression models \cite{skodras2001jpeg, webp, bpg, jpegxr} cannot be utilized on embedded MCUs due to their stark resource demands.
In contrast, \name employs a lightweight encoder that is executable on weak devices such as nRF5340 (Coretex M33) and STM32F7 (Cortex-M7).

\noindent
\textbf{Progressive image compression:} Most classic compression models offer a progressive option \cite{skodras2001jpeg, webp, bpg, jpegxr}. Among them, ProgJPEG \cite{wallace1991jpeg} is the only one that can be deployed on resource-constrained embedded devices. 
Recently, learning-based compression models \cite{ye2023accelir, diao2020drasic, su2020scalable, lee2018context, liu2021swin} have emerged, employing diverse techniques such as RNNs \cite{toderici2017full, toderici2015variable, gregor2016towards, johnston2018improved}, bit prioritization \cite{jeon2023context, lee2022dpict}, multiple decoders \cite{cai2019novel}, and nested quantization \cite{lu2021progressive} to produce progressive bitstreams. Similar to their non-progressive counterparts, these deep models are unsuitable for weak devices due to their high resource demands. 

\begin{table}[tb]
\caption{\note{Top-1 accuracy of \name, ProgJPEG, and Starfish on CIFAR100 with 2.5KB/s bandwidth and different Packet Loss Rates (PLR) after a single connection cycle (0.74-sec of transmission per minute). The first number shows the loss rate, and the second shows the received bytes after a cycle.}}
\vspace{-1em}
\begin{tabular}{|l|l|l|l|l|}
\hline
& \textbf{ 10\% PLR} & \textbf{ 40\% PLR} & \textbf{ 70\% PLR}\\
& \textbf{ (1.62 KB)} & \textbf{ (1.31 KB)} & \textbf{ (0.5 KB)}\\
\hline
\textbf{\name} & \textbf{82.06\%} & \textbf{81.2\%}  & \textbf{71.4\%}\\
ProgJPEG & 61.7\% & 43.1\% & 0\%\\
Starfish & 10.2\% & 2.6\% & 1\%\\
\hline
\end{tabular}
\label{tab:packet_loss}
\vspace{-1em}
\end{table}

\noindent
\textbf{Edge offloading models: } We categorize offloading models into two types: those with autoencoders \cite{matsubara2019distilled, hu2020fast, jankowski2020joint, matsubara2020head, yao2020deep, shao2020bottlenet++, rethinking, huang2022real, huang2020clio}, and those without autoencoders \cite{zeng2019boomerang, pagliari2020crime, li2018learning, li2018auto, jeong2018computation, eshratifar2019jointdnn, chen2021deep}. While these models are usually focused on vision tasks, other models explore offloading for further tasks, including speech recognition and semantic segmentation \cite{choi2020back, matsubara2021neural, matsubara2020split, assine2021single, lee2021splittable, matsubara2022supervised, matsubara2022sc2}. However, in contrast to \name, all of these models lack progressive offloading.

\noindent
\textbf{Importance Detection: } We categorize importance detection models into three types: ROI detection \cite{ren2015faster, redmon2016you, he2017mask, uijlings2013selective, dai2016r}, Explainable-AI \cite{selvaraju2017gradcam, chattopadhyay2018gradcam++, wang2020scorecam, fu2020axiombased}, and Saliency Detection \cite{liu2018picanet, wang2019bilateral, wang2018recurrent, zhang2018nested, hou2017deeply, li2018amulet}. However, running these models on embedded devices is often infeasible due to their high computational and resource costs, comparable to or even greater than running a full classifier. In \name we use a lightweight saliency detector, which is not as accurate as these models but can significantly improve the performance of progressive offloading when dealing with limited bandwidth.

\section{Discussion}

\noindent
\textbf{Limitations: } \name generates a progressive bitstream, but this characteristic comes with a drawback as illustrated in Fig.~\ref{accuracy}. The progressive nature of the bitstream hinders \name from achieving its maximum performance. The reason is that when only part of the data is available for decoding, the model expends effort in handling this incompleteness, degrading the maximum accuracy attainable.

\noindent
\textbf{ImageNet1000/COCO performance: }
As reported in Fig.~\ref{fig:ACC-ImageNet} and Fig.~\ref{fig:mAP-COCO},  the performance of \name on ImageNet1000 and COCO is not as good as state-of-the-art. Nevertheless, we evaluated our model on ImageNet1000 to show the maximum capability. However, for real-world applications on weak devices, such as in the recent literature \cite{huang2022real, rethinking, hu2020starfish, huang2020clio}, the target task is usually much simpler, e.g., similar to CIFAR100. \review{Furthermore, ImageNet1000's image sizes are too large ($224 \times 224$ after preprocessing), each occupying around 150KB, which is a significant memory overhead for MCUs. Therefore, for such constrained devices, it is reasonable to opt for smaller input sizes.} 

% \noindent
% \review{\textbf{Other vision tasks:} \name incorporates a saliency detection mechanism. Therefore, it works on any vision task that relies mainly on the foreground of an image.
% Given \name's superior performance on CIFAR100 compared to ImageNet1000, we believe that \name can yield promising results on less complex datasets like MNIST \cite{lecun2010mnist}, SVHN \cite{netzer2011svhn} and Flowers \cite{nilsback2008automated}. On the other hand, our observations indicate that \name's performance diminishes on more challenging tasks such as object detection when precise focus on multiple objects is required, see Fig.~\ref{accuracy}. 

% }

\noindent
\review{
\textbf{Comparison to the SOTA:} Starfish \cite{hu2020starfish} uses dropout to add redundancy, which can deal with packet loss and is a first step towards progressiveness, as its accuracy does not collapse when receiving partial data. However, it is important to guide the model towards prioritizing the main objects, using saliency supported by gradual scoring (see Fig. \ref{fig:demo}) when transmitting. Starfish and ProgJPEG fail to do so, resulting in inferior performance (see Fig. \ref{accuracy}). FLEET \cite{rethinking}, on the other hand, has 4 fixed progressive steps, transmitting at least 2 KB of values, in each step. In contrast to the existing SOTA, \name can progressively transmit the encoded data value by value in the order of importance, after initially transmitting the saliency map (at most 40 bytes) which results in a much more granular and flexible progressiveness.
}

% \noindent
% \review{
% \textbf{Limitations of other alternatives:}
% Pruning and quantization techniques fail to generate MCU-compatible models due to substantial performance degradation when pruning below 10\%. Also, existing SOTA image compression models such as Ballé are too big for MCUs (165 times more parameters).
% Furthermore, despite the good performance of on-device inference models \cite{lin2020mcunet, lin2021mcunetv2}, they are designed for specific network architecture and are not generalizable. Therefore, running a larger and better classifier on the cloud can be beneficial to achieve a higher accuracy. Additionally, with our design, it is possible to exchange the classifier on the server to either use an even better model or a model for a different task without needing to change anything on the MCU.
% }

\section{Conclusion}

In this paper, we introduce \name, an image compression and offloading model designed for extremely weak devices under LPWANs limitations: low bandwidth, short duty cycles, and high packet loss rates. We address these challenges by developing a lightweight content-aware progressive encoding scheme that prioritizes critical data during transmission based on their relative effects on classification accuracy. This progressive offloading allows the cloud to run the inference even with partial data availability, which is crucial when we have time-sensitive inferences such as deadlines or limited duty cycle budgets.

\begin{acks}
\label{sec:acknowledgements}
\review{We thank our anonymous reviewers and our shepherd, Yuanchun Li, for their insight and suggestions for improvement.} This project has received funding from the Federal Ministry for Digital and Transport under the CAPTN-Fj\"{o}rd 5G project grant no.~45FGU139\_H , Federal Ministry for Economic Affairs and Climate Action under the Marispace-X project grant no.~68GX21002 and Marine Data Science (MarDATA) grant no.~HIDSS-0005.

\end{acks}

\bibliographystyle{ACM-Reference-Format}
\bibliography{sample-base}

%%% -*-BibTeX-*-
%%% Do NOT edit. File created by BibTeX with style
%%% ACM-Reference-Format-Journals [18-Jan-2012].

\begin{thebibliography}{102}

%%% ====================================================================
%%% NOTE TO THE USER: you can override these defaults by providing
%%% customized versions of any of these macros before the \bibliography
%%% command.  Each of them MUST provide its own final punctuation,
%%% except for \shownote{}, \showDOI{}, and \showURL{}.  The latter two
%%% do not use final punctuation, in order to avoid confusing it with
%%% the Web address.
%%%
%%% To suppress output of a particular field, define its macro to expand
%%% to an empty string, or better, \unskip, like this:
%%%
%%% \newcommand{\showDOI}[1]{\unskip}   % LaTeX syntax
%%%
%%% \def \showDOI #1{\unskip}           % plain TeX syntax
%%%
%%% ====================================================================

\ifx \showCODEN    \undefined \def \showCODEN     #1{\unskip}     \fi
\ifx \showDOI      \undefined \def \showDOI       #1{#1}\fi
\ifx \showISBNx    \undefined \def \showISBNx     #1{\unskip}     \fi
\ifx \showISBNxiii \undefined \def \showISBNxiii  #1{\unskip}     \fi
\ifx \showISSN     \undefined \def \showISSN      #1{\unskip}     \fi
\ifx \showLCCN     \undefined \def \showLCCN      #1{\unskip}     \fi
\ifx \shownote     \undefined \def \shownote      #1{#1}          \fi
\ifx \showarticletitle \undefined \def \showarticletitle #1{#1}   \fi
\ifx \showURL      \undefined \def \showURL       {\relax}        \fi
% The following commands are used for tagged output and should be
% invisible to TeX
\providecommand\bibfield[2]{#2}
\providecommand\bibinfo[2]{#2}
\providecommand\natexlab[1]{#1}
\providecommand\showeprint[2][]{arXiv:#2}

\bibitem[git({[n.\,d.]})]%
        {github:JPEGEncoder4Cortex-M}
 \bibinfo{year}{[n.\,d.]}\natexlab{}.
\newblock \bibinfo{title}{{JPEGEncoder4Cortex}}.
\newblock
  \bibinfo{howpublished}{\url{https://github.com/noritsuna/JPEGEncoder4Cortex-M/tree/master}}.
\newblock


\bibitem[tfl({[n.\,d.]})]%
        {tflite-micro}
 \bibinfo{year}{[n.\,d.]}\natexlab{}.
\newblock \bibinfo{title}{{TensorFlow Lite Micro}}.
\newblock \bibinfo{howpublished}{\url{https://www.tensorflow.org/lite/micro}}.
\newblock
\newblock
\shownote{Accessed: [2023]}.


\bibitem[zep({[n.\,d.]})]%
        {zephyr}
 \bibinfo{year}{[n.\,d.]}\natexlab{}.
\newblock \bibinfo{title}{{Zephyr Project RTOS}}.
\newblock
  \bibinfo{howpublished}{\url{https://github.com/zephyrproject-rtos/zephyr}}.
\newblock
\newblock
\shownote{Accessed: [2023]}.


\bibitem[jpe(2009)]%
        {jpegxr}
 \bibinfo{year}{2009}\natexlab{}.
\newblock \bibinfo{title}{{JPEG} {XR} Specification}.
\newblock
\newblock
\urldef\tempurl%
\url{https://jpeg.org/jpegxr/}
\showURL{%
\tempurl}
\newblock
\shownote{Joint Photographic Experts Group ({JPEG})}.


\bibitem[nRF(2023)]%
        {nRF5340}
 \bibinfo{year}{2023}\natexlab{}.
\newblock \bibinfo{booktitle}{\emph{nRF5340 System-on-Chip}}.
\newblock
\urldef\tempurl%
\url{https://www.nordicsemi.com/products/nrf5340}
\showURL{%
\tempurl}


\bibitem[STM(2023)]%
        {STM32F7Series}
 \bibinfo{year}{2023}\natexlab{}.
\newblock \bibinfo{booktitle}{\emph{STM32F7 Series Microcontrollers}}.
\newblock
\urldef\tempurl%
\url{https://www.st.com/en/microcontrollers-microprocessors/stm32f7-series.html}
\showURL{%
\tempurl}


\bibitem[Almeida et~al\mbox{.}(2022)]%
        {almeida2022dyno}
\bibfield{author}{\bibinfo{person}{Mario Almeida}, \bibinfo{person}{Stefanos
  Laskaridis}, \bibinfo{person}{Stylianos~I Venieris}, \bibinfo{person}{Ilias
  Leontiadis}, {and} \bibinfo{person}{Nicholas~D Lane}.}
  \bibinfo{year}{2022}\natexlab{}.
\newblock \showarticletitle{Dyno: Dynamic onloading of deep neural networks
  from cloud to device}.
\newblock \bibinfo{journal}{\emph{ACM Transactions on Embedded Computing
  Systems}} \bibinfo{volume}{21}, \bibinfo{number}{6} (\bibinfo{year}{2022}),
  \bibinfo{pages}{1--24}.
\newblock


\bibitem[{Apple Inc.}(nd)]%
        {apple_emergency_sos}
\bibfield{author}{\bibinfo{person}{{Apple Inc.}}}
  \bibinfo{year}{n.d.}\natexlab{}.
\newblock \bibinfo{booktitle}{\emph{Use Emergency SOS via satellite on your
  iPhone}}.
\newblock
\urldef\tempurl%
\url{https://support.apple.com/en-us/HT213426}
\showURL{%
\tempurl}
\newblock
\shownote{Accessed: November 11, 2023}.


\bibitem[Assine et~al\mbox{.}(2021)]%
        {assine2021single}
\bibfield{author}{\bibinfo{person}{Juliano~S Assine}, \bibinfo{person}{Eduardo
  Valle}, {et~al\mbox{.}}} \bibinfo{year}{2021}\natexlab{}.
\newblock \showarticletitle{Single-training collaborative object detectors
  adaptive to bandwidth and computation}.
\newblock \bibinfo{journal}{\emph{arXiv preprint arXiv:2105.00591}}
  (\bibinfo{year}{2021}).
\newblock


\bibitem[Avazov et~al\mbox{.}(2023)]%
        {avazov2023forest}
\bibfield{author}{\bibinfo{person}{Kuldoshbay Avazov}, \bibinfo{person}{An~Eui
  Hyun}, \bibinfo{person}{Alabdulwahab~Abrar Sami~S}, \bibinfo{person}{Azizbek
  Khaitov}, \bibinfo{person}{Akmalbek~Bobomirzaevich Abdusalomov}, {and}
  \bibinfo{person}{Young~Im Cho}.} \bibinfo{year}{2023}\natexlab{}.
\newblock \showarticletitle{Forest Fire Detection and Notification Method Based
  on AI and IoT Approaches}.
\newblock \bibinfo{journal}{\emph{Future Internet}} \bibinfo{volume}{15},
  \bibinfo{number}{2} (\bibinfo{year}{2023}), \bibinfo{pages}{61}.
\newblock


\bibitem[Ball{\'e} et~al\mbox{.}(2015)]%
        {balle2015density}
\bibfield{author}{\bibinfo{person}{Johannes Ball{\'e}}, \bibinfo{person}{Valero
  Laparra}, {and} \bibinfo{person}{Eero~P Simoncelli}.}
  \bibinfo{year}{2015}\natexlab{}.
\newblock \showarticletitle{Density modeling of images using a generalized
  normalization transformation}.
\newblock \bibinfo{journal}{\emph{arXiv preprint arXiv:1511.06281}}
  (\bibinfo{year}{2015}).
\newblock


\bibitem[Ball{\'e} et~al\mbox{.}(2016)]%
        {Balle2016end}
\bibfield{author}{\bibinfo{person}{Johannes Ball{\'e}}, \bibinfo{person}{Valero
  Laparra}, {and} \bibinfo{person}{Eero~P Simoncelli}.}
  \bibinfo{year}{2016}\natexlab{}.
\newblock \showarticletitle{End-to-end optimized image compression}.
\newblock \bibinfo{journal}{\emph{arXiv preprint arXiv:1611.01704}}
  (\bibinfo{year}{2016}).
\newblock


\bibitem[Ball{\'e} et~al\mbox{.}(2018)]%
        {balle2018variational}
\bibfield{author}{\bibinfo{person}{Johannes Ball{\'e}}, \bibinfo{person}{David
  Minnen}, \bibinfo{person}{Saurabh Singh}, \bibinfo{person}{Sung~Jin Hwang},
  {and} \bibinfo{person}{Nick Johnston}.} \bibinfo{year}{2018}\natexlab{}.
\newblock \showarticletitle{Variational image compression with a scale
  hyperprior}.
\newblock \bibinfo{journal}{\emph{arXiv preprint arXiv:1802.01436}}
  (\bibinfo{year}{2018}).
\newblock


\bibitem[Bj{\o}tegaard(2001)]%
        {bjotegaard2001calculation}
\bibfield{author}{\bibinfo{person}{G Bj{\o}tegaard}.}
  \bibinfo{year}{2001}\natexlab{}.
\newblock \showarticletitle{Calculation of average PSNR differences between
  RD-curves (vceg-m33)}. In \bibinfo{booktitle}{\emph{VCEG Meeting (ITU-T SG16
  Q. 6), Austin, Texas, USA,, Tech. Rep. M}}, Vol.~\bibinfo{volume}{16090}.
\newblock


\bibitem[Bor et~al\mbox{.}(2016)]%
        {bor2016lora}
\bibfield{author}{\bibinfo{person}{Martin Bor}, \bibinfo{person}{John~Edward
  Vidler}, {and} \bibinfo{person}{Utz Roedig}.}
  \bibinfo{year}{2016}\natexlab{}.
\newblock \showarticletitle{LoRa for the Internet of Things}.
\newblock  (\bibinfo{year}{2016}).
\newblock


\bibitem[BPG({[n.\,d.]})]%
        {bpg}
BPG \bibinfo{year}{[n.\,d.]}\natexlab{}.
\newblock \bibinfo{title}{BPG Image format}.
\newblock \bibinfo{howpublished}{\url{https://bellard.org/bpg/}}.
\newblock
\newblock
\shownote{Accessed: 2023-02-14}.


\bibitem[Cai et~al\mbox{.}(2019a)]%
        {cai2019end}
\bibfield{author}{\bibinfo{person}{Chunlei Cai}, \bibinfo{person}{Li Chen},
  \bibinfo{person}{Xiaoyun Zhang}, {and} \bibinfo{person}{Zhiyong Gao}.}
  \bibinfo{year}{2019}\natexlab{a}.
\newblock \showarticletitle{End-to-end optimized ROI image compression}.
\newblock \bibinfo{journal}{\emph{IEEE Transactions on Image Processing}}
  \bibinfo{volume}{29} (\bibinfo{year}{2019}), \bibinfo{pages}{3442--3457}.
\newblock


\bibitem[Cai et~al\mbox{.}(2019b)]%
        {cai2019novel}
\bibfield{author}{\bibinfo{person}{Chunlei Cai}, \bibinfo{person}{Li Chen},
  \bibinfo{person}{Xiaoyun Zhang}, \bibinfo{person}{Guo Lu}, {and}
  \bibinfo{person}{Zhiyong Gao}.} \bibinfo{year}{2019}\natexlab{b}.
\newblock \showarticletitle{A novel deep progressive image compression
  framework}. In \bibinfo{booktitle}{\emph{2019 Picture Coding Symposium
  (PCS)}}. IEEE, \bibinfo{pages}{1--5}.
\newblock


\bibitem[Camal and Aksanli(2020)]%
        {camal2020building}
\bibfield{author}{\bibinfo{person}{Luis Camal} {and} \bibinfo{person}{Baris
  Aksanli}.} \bibinfo{year}{2020}\natexlab{}.
\newblock \showarticletitle{Building an energy-efficient ad-hoc network for
  wildlife observation}.
\newblock \bibinfo{journal}{\emph{Electronics}} \bibinfo{volume}{9},
  \bibinfo{number}{6} (\bibinfo{year}{2020}), \bibinfo{pages}{984}.
\newblock


\bibitem[Chattopadhyay et~al\mbox{.}(2020)]%
        {chattopadhyay2018gradcam++}
\bibfield{author}{\bibinfo{person}{Aditya Chattopadhyay},
  \bibinfo{person}{Anirban Sarkar}, \bibinfo{person}{Prantik Howlader}, {and}
  \bibinfo{person}{Vineeth~N Balasubramanian}.}
  \bibinfo{year}{2020}\natexlab{}.
\newblock \showarticletitle{Grad-CAM++: Improved Visual Explanations for Deep
  Convolutional Networks}.
\newblock \bibinfo{journal}{\emph{IEEE Transactions on Image Processing}}
  \bibinfo{volume}{29} (\bibinfo{year}{2020}), \bibinfo{pages}{4030--4043}.
\newblock


\bibitem[Chaudhari et~al\mbox{.}(2020)]%
        {chaudhari2020lpwan}
\bibfield{author}{\bibinfo{person}{Bharat~S Chaudhari}, \bibinfo{person}{Marco
  Zennaro}, {and} \bibinfo{person}{Suresh Borkar}.}
  \bibinfo{year}{2020}\natexlab{}.
\newblock \showarticletitle{LPWAN technologies: Emerging application
  characteristics, requirements, and design considerations}.
\newblock \bibinfo{journal}{\emph{Future Internet}} \bibinfo{volume}{12},
  \bibinfo{number}{3} (\bibinfo{year}{2020}), \bibinfo{pages}{46}.
\newblock


\bibitem[Chen et~al\mbox{.}(2021)]%
        {chen2021deep}
\bibfield{author}{\bibinfo{person}{Bo Chen}, \bibinfo{person}{Zhisheng Yan},
  \bibinfo{person}{Hongpeng Guo}, \bibinfo{person}{Zhe Yang},
  \bibinfo{person}{Ahmed Ali-Eldin}, \bibinfo{person}{Prashant Shenoy}, {and}
  \bibinfo{person}{Klara Nahrstedt}.} \bibinfo{year}{2021}\natexlab{}.
\newblock \showarticletitle{Deep contextualized compressive offloading for
  images}. In \bibinfo{booktitle}{\emph{Proceedings of the 19th ACM Conference
  on Embedded Networked Sensor Systems}}. \bibinfo{pages}{467--473}.
\newblock


\bibitem[Cheng et~al\mbox{.}(2020)]%
        {cheng2020learned}
\bibfield{author}{\bibinfo{person}{Zhengxue Cheng}, \bibinfo{person}{Heming
  Sun}, \bibinfo{person}{Masaru Takeuchi}, {and} \bibinfo{person}{Jiro Katto}.}
  \bibinfo{year}{2020}\natexlab{}.
\newblock \showarticletitle{Learned image compression with discretized gaussian
  mixture likelihoods and attention modules}. In
  \bibinfo{booktitle}{\emph{Proceedings of the IEEE/CVF Conference on Computer
  Vision and Pattern Recognition}}. \bibinfo{pages}{7939--7948}.
\newblock


\bibitem[Choi et~al\mbox{.}(2020)]%
        {choi2020back}
\bibfield{author}{\bibinfo{person}{Hyomin Choi}, \bibinfo{person}{Robert~A
  Cohen}, {and} \bibinfo{person}{Ivan~V Baji{\'c}}.}
  \bibinfo{year}{2020}\natexlab{}.
\newblock \showarticletitle{Back-and-forth prediction for deep tensor
  compression}. In \bibinfo{booktitle}{\emph{ICASSP 2020-2020 IEEE
  International Conference on Acoustics, Speech and Signal Processing
  (ICASSP)}}. IEEE, \bibinfo{pages}{4467--4471}.
\newblock


\bibitem[Choi et~al\mbox{.}(2019)]%
        {choi2019latency}
\bibfield{author}{\bibinfo{person}{HeeSeok Choi}, \bibinfo{person}{Heonchang
  Yu}, {and} \bibinfo{person}{EunYoung Lee}.} \bibinfo{year}{2019}\natexlab{}.
\newblock \showarticletitle{Latency-classification-based deadline-aware task
  offloading algorithm in mobile edge computing environments}.
\newblock \bibinfo{journal}{\emph{Applied Sciences}} \bibinfo{volume}{9},
  \bibinfo{number}{21} (\bibinfo{year}{2019}), \bibinfo{pages}{4696}.
\newblock


\bibitem[Dai et~al\mbox{.}(2016)]%
        {dai2016r}
\bibfield{author}{\bibinfo{person}{Jifeng Dai}, \bibinfo{person}{Yi Li},
  \bibinfo{person}{Kaiming He}, {and} \bibinfo{person}{Jian Sun}.}
  \bibinfo{year}{2016}\natexlab{}.
\newblock \showarticletitle{{R-FCN}: Object Detection via Region-based Fully
  Convolutional Networks}. In \bibinfo{booktitle}{\emph{Advances in Neural
  Information Processing Systems}}. \bibinfo{pages}{379--387}.
\newblock


\bibitem[Deng et~al\mbox{.}(2009)]%
        {deng2009imagenet}
\bibfield{author}{\bibinfo{person}{Jia Deng}, \bibinfo{person}{Wei Dong},
  \bibinfo{person}{Richard Socher}, \bibinfo{person}{Li-Jia Li},
  \bibinfo{person}{Kai Li}, {and} \bibinfo{person}{Li Fei-Fei}.}
  \bibinfo{year}{2009}\natexlab{}.
\newblock \showarticletitle{Imagenet: A large-scale hierarchical image
  database}. In \bibinfo{booktitle}{\emph{2009 IEEE conference on computer
  vision and pattern recognition}}. Ieee, \bibinfo{pages}{248--255}.
\newblock


\bibitem[Diao et~al\mbox{.}(2020)]%
        {diao2020drasic}
\bibfield{author}{\bibinfo{person}{Enmao Diao}, \bibinfo{person}{Jie Ding},
  {and} \bibinfo{person}{Vahid Tarokh}.} \bibinfo{year}{2020}\natexlab{}.
\newblock \showarticletitle{Drasic: Distributed recurrent autoencoder for
  scalable image compression}. In \bibinfo{booktitle}{\emph{2020 Data
  Compression Conference (DCC)}}. IEEE, \bibinfo{pages}{3--12}.
\newblock


\bibitem[Dosovitskiy et~al\mbox{.}(2021)]%
        {dosovitskiy2020image}
\bibfield{author}{\bibinfo{person}{Alexey Dosovitskiy}, \bibinfo{person}{Lucas
  Beyer}, \bibinfo{person}{Alexander Kolesnikov}, \bibinfo{person}{Dirk
  Weissenborn}, \bibinfo{person}{Xiaohua Zhai}, \bibinfo{person}{Thomas
  Unterthiner}, \bibinfo{person}{Mostafa Dehghani}, \bibinfo{person}{Matthias
  Minderer}, \bibinfo{person}{Georg Heigold}, \bibinfo{person}{Sylvain Gelly},
  \bibinfo{person}{Jakob Uszkoreit}, {and} \bibinfo{person}{Neil Houlsby}.}
  \bibinfo{year}{2021}\natexlab{}.
\newblock \showarticletitle{An image is worth 16x16 words: Transformers for
  image recognition at scale}. In \bibinfo{booktitle}{\emph{Proceedings of the
  IEEE/CVF Conference on Computer Vision and Pattern Recognition}}.
  \bibinfo{pages}{3156--3164}.
\newblock


\bibitem[Eshratifar et~al\mbox{.}(2019a)]%
        {eshratifar2019jointdnn}
\bibfield{author}{\bibinfo{person}{Amir~Erfan Eshratifar},
  \bibinfo{person}{Mohammad~Saeed Abrishami}, {and} \bibinfo{person}{Massoud
  Pedram}.} \bibinfo{year}{2019}\natexlab{a}.
\newblock \showarticletitle{JointDNN: An efficient training and inference
  engine for intelligent mobile cloud computing services}.
\newblock \bibinfo{journal}{\emph{IEEE Transactions on Mobile Computing}}
  \bibinfo{volume}{20}, \bibinfo{number}{2} (\bibinfo{year}{2019}),
  \bibinfo{pages}{565--576}.
\newblock


\bibitem[Eshratifar et~al\mbox{.}(2019b)]%
        {eshratifar2019bottlenet}
\bibfield{author}{\bibinfo{person}{Amir~Erfan Eshratifar},
  \bibinfo{person}{Amirhossein Esmaili}, {and} \bibinfo{person}{Massoud
  Pedram}.} \bibinfo{year}{2019}\natexlab{b}.
\newblock \showarticletitle{Bottlenet: A deep learning architecture for
  intelligent mobile cloud computing services}. In
  \bibinfo{booktitle}{\emph{2019 IEEE/ACM International Symposium on Low Power
  Electronics and Design (ISLPED)}}. IEEE, \bibinfo{pages}{1--6}.
\newblock


\bibitem[Fourati and Alouini(2021)]%
        {fourati2021artificial}
\bibfield{author}{\bibinfo{person}{Fares Fourati} {and}
  \bibinfo{person}{Mohamed-Slim Alouini}.} \bibinfo{year}{2021}\natexlab{}.
\newblock \showarticletitle{Artificial intelligence for satellite
  communication: A review}.
\newblock \bibinfo{journal}{\emph{Intelligent and Converged Networks}}
  \bibinfo{volume}{2}, \bibinfo{number}{3} (\bibinfo{year}{2021}),
  \bibinfo{pages}{213--243}.
\newblock


\bibitem[Fu et~al\mbox{.}(2020)]%
        {fu2020axiombased}
\bibfield{author}{\bibinfo{person}{Ruigang Fu}, \bibinfo{person}{Qingyong Hu},
  \bibinfo{person}{Xiaohu Dong}, \bibinfo{person}{Yulan Guo},
  \bibinfo{person}{Yinghui Gao}, {and} \bibinfo{person}{Biao Li}.}
  \bibinfo{year}{2020}\natexlab{}.
\newblock \showarticletitle{Axiom-based Grad-CAM: Towards Accurate
  Visualization and Explanation of CNNs}.
\newblock \bibinfo{journal}{\emph{arXiv preprint arXiv:2008.02312}}
  (\bibinfo{year}{2020}).
\newblock


\bibitem[Gregor et~al\mbox{.}(2016)]%
        {gregor2016towards}
\bibfield{author}{\bibinfo{person}{Karol Gregor}, \bibinfo{person}{Frederic
  Besse}, \bibinfo{person}{Danilo Jimenez~Rezende}, \bibinfo{person}{Ivo
  Danihelka}, {and} \bibinfo{person}{Daan Wierstra}.}
  \bibinfo{year}{2016}\natexlab{}.
\newblock \showarticletitle{Towards conceptual compression}.
\newblock \bibinfo{journal}{\emph{Advances In Neural Information Processing
  Systems}}  \bibinfo{volume}{29} (\bibinfo{year}{2016}).
\newblock


\bibitem[He et~al\mbox{.}(2017)]%
        {he2017mask}
\bibfield{author}{\bibinfo{person}{Kaiming He}, \bibinfo{person}{Georgia
  Gkioxari}, \bibinfo{person}{Piotr Dollar}, {and} \bibinfo{person}{Ross
  Girshick}.} \bibinfo{year}{2017}\natexlab{}.
\newblock \showarticletitle{Mask {R-CNN}}. In
  \bibinfo{booktitle}{\emph{Proceedings of the IEEE International Conference on
  Computer Vision}}. IEEE, \bibinfo{pages}{2961--2969}.
\newblock


\bibitem[He et~al\mbox{.}(2016)]%
        {he2016deep}
\bibfield{author}{\bibinfo{person}{Kaiming He}, \bibinfo{person}{Xiangyu
  Zhang}, \bibinfo{person}{Shaoqing Ren}, {and} \bibinfo{person}{Jian Sun}.}
  \bibinfo{year}{2016}\natexlab{}.
\newblock \showarticletitle{Deep residual learning for image recognition}. In
  \bibinfo{booktitle}{\emph{Proceedings of the IEEE conference on computer
  vision and pattern recognition}}. \bibinfo{pages}{770--778}.
\newblock


\bibitem[Hinton et~al\mbox{.}(2015)]%
        {hinton2015distilling}
\bibfield{author}{\bibinfo{person}{Geoffrey Hinton}, \bibinfo{person}{Oriol
  Vinyals}, {and} \bibinfo{person}{Jeff Dean}.}
  \bibinfo{year}{2015}\natexlab{}.
\newblock \showarticletitle{Distilling the knowledge in a neural network}.
\newblock \bibinfo{journal}{\emph{arXiv preprint arXiv:1503.02531}}
  (\bibinfo{year}{2015}).
\newblock


\bibitem[Hojjat et~al\mbox{.}(2023)]%
        {Hojjat_2023_CVPR}
\bibfield{author}{\bibinfo{person}{Ali Hojjat}, \bibinfo{person}{Janek
  Haberer}, {and} \bibinfo{person}{Olaf Landsiedel}.}
  \bibinfo{year}{2023}\natexlab{}.
\newblock \showarticletitle{ProgDTD: Progressive Learned Image Compression With
  Double-Tail-Drop Training}. In \bibinfo{booktitle}{\emph{Proceedings of the
  IEEE/CVF Conference on Computer Vision and Pattern Recognition (CVPR)
  Workshops}}. \bibinfo{pages}{1130--1139}.
\newblock


\bibitem[Hou et~al\mbox{.}(2017)]%
        {hou2017deeply}
\bibfield{author}{\bibinfo{person}{Qibin Hou}, \bibinfo{person}{Ming-Ming
  Cheng}, \bibinfo{person}{Xiaowei Hu}, \bibinfo{person}{Ali Borji},
  \bibinfo{person}{Zhuowen Tu}, {and} \bibinfo{person}{Philip Torr}.}
  \bibinfo{year}{2017}\natexlab{}.
\newblock \showarticletitle{Deeply Supervised Salient Object Detection with
  Short Connections}. In \bibinfo{booktitle}{\emph{Proceedings of the IEEE
  Conference on Computer Vision and Pattern Recognition (CVPR)}}.
\newblock


\bibitem[Hu and Krishnamachari(2020)]%
        {hu2020fast}
\bibfield{author}{\bibinfo{person}{Diyi Hu} {and} \bibinfo{person}{Bhaskar
  Krishnamachari}.} \bibinfo{year}{2020}\natexlab{}.
\newblock \showarticletitle{Fast and accurate streaming CNN inference via
  communication compression on the edge}. In \bibinfo{booktitle}{\emph{2020
  IEEE/ACM Fifth International Conference on Internet-of-Things Design and
  Implementation (IoTDI)}}. IEEE, \bibinfo{pages}{157--163}.
\newblock


\bibitem[Hu et~al\mbox{.}(2020)]%
        {hu2020starfish}
\bibfield{author}{\bibinfo{person}{Pan Hu}, \bibinfo{person}{Junha Im},
  \bibinfo{person}{Zain Asgar}, {and} \bibinfo{person}{Sachin Katti}.}
  \bibinfo{year}{2020}\natexlab{}.
\newblock \showarticletitle{Starfish: resilient image compression for aiot
  cameras}. In \bibinfo{booktitle}{\emph{Proceedings of the 18th Conference on
  Embedded Networked Sensor Systems}}. \bibinfo{pages}{395--408}.
\newblock


\bibitem[Huang et~al\mbox{.}(2023)]%
        {rethinking}
\bibfield{author}{\bibinfo{person}{Jin Huang}, \bibinfo{person}{Hui Guan},
  {and} \bibinfo{person}{Deepak Ganesan}.} \bibinfo{year}{2023}\natexlab{}.
\newblock \showarticletitle{Re-Thinking Computation Offload for Efficient
  Inference on IoT Devices with Duty-Cycled Radios}. In
  \bibinfo{booktitle}{\emph{Proceedings of the 29th Annual International
  Conference on Mobile Computing and Networking}} (Madrid, Spain)
  \emph{(\bibinfo{series}{ACM MobiCom '23})}. \bibinfo{publisher}{Association
  for Computing Machinery}, \bibinfo{address}{New York, NY, USA}, Article
  \bibinfo{articleno}{13}, \bibinfo{numpages}{15}~pages.
\newblock
\showISBNx{9781450399906}
\urldef\tempurl%
\url{https://doi.org/10.1145/3570361.3592514}
\showDOI{\tempurl}


\bibitem[Huang et~al\mbox{.}(2020)]%
        {huang2020clio}
\bibfield{author}{\bibinfo{person}{Jin Huang}, \bibinfo{person}{Colin
  Samplawski}, \bibinfo{person}{Deepak Ganesan}, \bibinfo{person}{Benjamin
  Marlin}, {and} \bibinfo{person}{Heesung Kwon}.}
  \bibinfo{year}{2020}\natexlab{}.
\newblock \showarticletitle{Clio: Enabling automatic compilation of deep
  learning pipelines across iot and cloud}. In
  \bibinfo{booktitle}{\emph{Proceedings of the 26th Annual International
  Conference on Mobile Computing and Networking}}. \bibinfo{pages}{1--12}.
\newblock


\bibitem[Huang and Gao(2022)]%
        {huang2022real}
\bibfield{author}{\bibinfo{person}{Kai Huang} {and} \bibinfo{person}{Wei Gao}.}
  \bibinfo{year}{2022}\natexlab{}.
\newblock \showarticletitle{Real-time neural network inference on extremely
  weak devices: agile offloading with explainable AI}. In
  \bibinfo{booktitle}{\emph{Proceedings of the 28th Annual International
  Conference on Mobile Computing And Networking}}. \bibinfo{pages}{200--213}.
\newblock


\bibitem[Jankowski et~al\mbox{.}(2020)]%
        {jankowski2020joint}
\bibfield{author}{\bibinfo{person}{Mikolaj Jankowski}, \bibinfo{person}{Deniz
  G{\"u}nd{\"u}z}, {and} \bibinfo{person}{Krystian Mikolajczyk}.}
  \bibinfo{year}{2020}\natexlab{}.
\newblock \showarticletitle{Joint device-edge inference over wireless links
  with pruning}. In \bibinfo{booktitle}{\emph{2020 IEEE 21st international
  workshop on signal processing advances in wireless communications (SPAWC)}}.
  IEEE, \bibinfo{pages}{1--5}.
\newblock


\bibitem[Jeon et~al\mbox{.}(2023)]%
        {jeon2023context}
\bibfield{author}{\bibinfo{person}{Seungmin Jeon}, \bibinfo{person}{Kwang~Pyo
  Choi}, \bibinfo{person}{Youngo Park}, {and} \bibinfo{person}{Chang-Su Kim}.}
  \bibinfo{year}{2023}\natexlab{}.
\newblock \showarticletitle{Context-Based Trit-Plane Coding for Progressive
  Image Compression}. In \bibinfo{booktitle}{\emph{Proceedings of the IEEE/CVF
  Conference on Computer Vision and Pattern Recognition}}.
  \bibinfo{pages}{14348--14357}.
\newblock


\bibitem[Jeong et~al\mbox{.}(2018)]%
        {jeong2018computation}
\bibfield{author}{\bibinfo{person}{Hyuk-Jin Jeong}, \bibinfo{person}{InChang
  Jeong}, \bibinfo{person}{Hyeon-Jae Lee}, {and} \bibinfo{person}{Soo-Mook
  Moon}.} \bibinfo{year}{2018}\natexlab{}.
\newblock \showarticletitle{Computation offloading for machine learning web
  apps in the edge server environment}. In \bibinfo{booktitle}{\emph{2018 IEEE
  38th International Conference on Distributed Computing Systems (ICDCS)}}.
  IEEE, \bibinfo{pages}{1492--1499}.
\newblock


\bibitem[Johnston et~al\mbox{.}(2018)]%
        {johnston2018improved}
\bibfield{author}{\bibinfo{person}{Nick Johnston}, \bibinfo{person}{Damien
  Vincent}, \bibinfo{person}{David Minnen}, \bibinfo{person}{Michele Covell},
  \bibinfo{person}{Saurabh Singh}, \bibinfo{person}{Troy Chinen},
  \bibinfo{person}{Sung~Jin Hwang}, \bibinfo{person}{Joel Shor}, {and}
  \bibinfo{person}{George Toderici}.} \bibinfo{year}{2018}\natexlab{}.
\newblock \showarticletitle{Improved lossy image compression with priming and
  spatially adaptive bit rates for recurrent networks}. In
  \bibinfo{booktitle}{\emph{Proceedings of the IEEE Conference on Computer
  Vision and Pattern Recognition}}. \bibinfo{pages}{4385--4393}.
\newblock


\bibitem[Koike-Akino and Wang(2020)]%
        {koike2020stochastic}
\bibfield{author}{\bibinfo{person}{Toshiaki Koike-Akino} {and}
  \bibinfo{person}{Ye Wang}.} \bibinfo{year}{2020}\natexlab{}.
\newblock \showarticletitle{Stochastic bottleneck: Rateless auto-encoder for
  flexible dimensionality reduction}. In \bibinfo{booktitle}{\emph{2020 IEEE
  International Symposium on Information Theory (ISIT)}}. IEEE,
  \bibinfo{pages}{2735--2740}.
\newblock


\bibitem[Kramer(1991)]%
        {kramer1991nonlinear}
\bibfield{author}{\bibinfo{person}{Mark~A Kramer}.}
  \bibinfo{year}{1991}\natexlab{}.
\newblock \showarticletitle{Nonlinear principal component analysis using
  autoassociative neural networks}.
\newblock \bibinfo{journal}{\emph{AIChE journal}} \bibinfo{volume}{37},
  \bibinfo{number}{2} (\bibinfo{year}{1991}), \bibinfo{pages}{233--243}.
\newblock


\bibitem[{Krizhevsky}(2009)]%
        {cifar100}
\bibfield{author}{\bibinfo{person}{A. {Krizhevsky}}.}
  \bibinfo{year}{2009}\natexlab{}.
\newblock \bibinfo{booktitle}{\emph{CIFAR-100 (Canadian Institute for Advanced
  Research)}}.
\newblock \bibinfo{type}{{T}echnical {R}eport}.
  \bibinfo{institution}{University of Toronto}.
\newblock
\urldef\tempurl%
\url{https://www.cs.toronto.edu/~kriz/cifar.html}
\showURL{%
\tempurl}


\bibitem[Laskaridis et~al\mbox{.}(2020)]%
        {laskaridis2020spinn}
\bibfield{author}{\bibinfo{person}{Stefanos Laskaridis},
  \bibinfo{person}{Stylianos~I Venieris}, \bibinfo{person}{Mario Almeida},
  \bibinfo{person}{Ilias Leontiadis}, {and} \bibinfo{person}{Nicholas~D Lane}.}
  \bibinfo{year}{2020}\natexlab{}.
\newblock \showarticletitle{SPINN: synergistic progressive inference of neural
  networks over device and cloud}. In \bibinfo{booktitle}{\emph{Proceedings of
  the 26th annual international conference on mobile computing and
  networking}}. \bibinfo{pages}{1--15}.
\newblock


\bibitem[Lavric et~al\mbox{.}(2019)]%
        {lavric2019sigfox}
\bibfield{author}{\bibinfo{person}{Alexandru Lavric}, \bibinfo{person}{Adrian~I
  Petrariu}, {and} \bibinfo{person}{Valentin Popa}.}
  \bibinfo{year}{2019}\natexlab{}.
\newblock \showarticletitle{Sigfox communication protocol: The new era of
  iot?}. In \bibinfo{booktitle}{\emph{2019 international conference on sensing
  and instrumentation in IoT Era (ISSI)}}. IEEE, \bibinfo{pages}{1--4}.
\newblock


\bibitem[Lee et~al\mbox{.}(2018)]%
        {lee2018context}
\bibfield{author}{\bibinfo{person}{Jooyoung Lee}, \bibinfo{person}{Seunghyun
  Cho}, {and} \bibinfo{person}{Seung-Kwon Beack}.}
  \bibinfo{year}{2018}\natexlab{}.
\newblock \showarticletitle{Context-adaptive entropy model for end-to-end
  optimized image compression}.
\newblock \bibinfo{journal}{\emph{arXiv preprint arXiv:1809.10452}}
  (\bibinfo{year}{2018}).
\newblock


\bibitem[Lee et~al\mbox{.}(2021)]%
        {lee2021splittable}
\bibfield{author}{\bibinfo{person}{Joo~Chan Lee}, \bibinfo{person}{Yongwoo
  Kim}, \bibinfo{person}{SungTae Moon}, {and} \bibinfo{person}{Jong~Hwan Ko}.}
  \bibinfo{year}{2021}\natexlab{}.
\newblock \showarticletitle{A splittable dnn-based object detector for
  edge-cloud collaborative real-time video inference}. In
  \bibinfo{booktitle}{\emph{2021 17th IEEE International Conference on Advanced
  Video and Signal Based Surveillance (AVSS)}}. IEEE, \bibinfo{pages}{1--8}.
\newblock


\bibitem[Lee et~al\mbox{.}(2022)]%
        {lee2022dpict}
\bibfield{author}{\bibinfo{person}{Jae-Han Lee}, \bibinfo{person}{Seungmin
  Jeon}, \bibinfo{person}{Kwang~Pyo Choi}, \bibinfo{person}{Youngo Park}, {and}
  \bibinfo{person}{Chang-Su Kim}.} \bibinfo{year}{2022}\natexlab{}.
\newblock \showarticletitle{DPICT: Deep progressive image compression using
  trit-planes}. In \bibinfo{booktitle}{\emph{Proceedings of the IEEE/CVF
  Conference on Computer Vision and Pattern Recognition}}.
  \bibinfo{pages}{16113--16122}.
\newblock


\bibitem[Li et~al\mbox{.}(2018a)]%
        {li2018auto}
\bibfield{author}{\bibinfo{person}{Guangli Li}, \bibinfo{person}{Lei Liu},
  \bibinfo{person}{Xueying Wang}, \bibinfo{person}{Xiao Dong},
  \bibinfo{person}{Peng Zhao}, {and} \bibinfo{person}{Xiaobing Feng}.}
  \bibinfo{year}{2018}\natexlab{a}.
\newblock \showarticletitle{Auto-tuning neural network quantization framework
  for collaborative inference between the cloud and edge}. In
  \bibinfo{booktitle}{\emph{Artificial Neural Networks and Machine
  Learning--ICANN 2018: 27th International Conference on Artificial Neural
  Networks, Rhodes, Greece, October 4-7, 2018, Proceedings, Part I 27}}.
  Springer, \bibinfo{pages}{402--411}.
\newblock


\bibitem[Li et~al\mbox{.}(2018b)]%
        {li2018learning}
\bibfield{author}{\bibinfo{person}{He Li}, \bibinfo{person}{Kaoru Ota}, {and}
  \bibinfo{person}{Mianxiong Dong}.} \bibinfo{year}{2018}\natexlab{b}.
\newblock \showarticletitle{Learning IoT in edge: Deep learning for the
  Internet of Things with edge computing}.
\newblock \bibinfo{journal}{\emph{IEEE network}} \bibinfo{volume}{32},
  \bibinfo{number}{1} (\bibinfo{year}{2018}), \bibinfo{pages}{96--101}.
\newblock


\bibitem[Li et~al\mbox{.}(2018c)]%
        {li2018amulet}
\bibfield{author}{\bibinfo{person}{Jun Li}, \bibinfo{person}{Yuhua Wang},
  \bibinfo{person}{Xin Qi}, \bibinfo{person}{Xuming Hou}, {and}
  \bibinfo{person}{Ming-Ming Liu}.} \bibinfo{year}{2018}\natexlab{c}.
\newblock \showarticletitle{AMULET: Aggregated Multi-Level Feature U-Net for
  Salient Object Detection}. In \bibinfo{booktitle}{\emph{Proceedings of the
  European Conference on Computer Vision (ECCV)}}.
\newblock


\bibitem[Li et~al\mbox{.}(2020)]%
        {li2020efficient}
\bibfield{author}{\bibinfo{person}{Mu Li}, \bibinfo{person}{Kede Ma},
  \bibinfo{person}{Jane You}, \bibinfo{person}{David Zhang}, {and}
  \bibinfo{person}{Wangmeng Zuo}.} \bibinfo{year}{2020}\natexlab{}.
\newblock \showarticletitle{Efficient and effective context-based convolutional
  entropy modeling for image compression}.
\newblock \bibinfo{journal}{\emph{IEEE Transactions on Image Processing}}
  \bibinfo{volume}{29} (\bibinfo{year}{2020}), \bibinfo{pages}{5900--5911}.
\newblock


\bibitem[Li(2001)]%
        {li2001overview}
\bibfield{author}{\bibinfo{person}{Weiping Li}.}
  \bibinfo{year}{2001}\natexlab{}.
\newblock \showarticletitle{Overview of fine granularity scalability in MPEG-4
  video standard}.
\newblock \bibinfo{journal}{\emph{IEEE Transactions on circuits and systems for
  video technology}} \bibinfo{volume}{11}, \bibinfo{number}{3}
  (\bibinfo{year}{2001}), \bibinfo{pages}{301--317}.
\newblock


\bibitem[Lim et~al\mbox{.}(2022)]%
        {lim2022leqnet}
\bibfield{author}{\bibinfo{person}{Jongseong Lim}, \bibinfo{person}{Sunghun
  Jung}, \bibinfo{person}{Chan JeGal}, \bibinfo{person}{Gwanghoon Jung},
  \bibinfo{person}{Jung~Ho Yoo}, \bibinfo{person}{Jin~Kyu Gahm}, {and}
  \bibinfo{person}{Giltae Song}.} \bibinfo{year}{2022}\natexlab{}.
\newblock \showarticletitle{LEQNet: Light Earthquake Deep Neural Network for
  Earthquake Detection and Phase Picking}.
\newblock \bibinfo{journal}{\emph{Frontiers in Earth Science}}
  \bibinfo{volume}{10} (\bibinfo{year}{2022}), \bibinfo{pages}{848237}.
\newblock


\bibitem[Lin et~al\mbox{.}(2014)]%
        {cocodataset}
\bibfield{author}{\bibinfo{person}{Tsung{-}Yi Lin}, \bibinfo{person}{Michael
  Maire}, \bibinfo{person}{Serge~J. Belongie}, \bibinfo{person}{Lubomir~D.
  Bourdev}, \bibinfo{person}{Ross~B. Girshick}, \bibinfo{person}{James Hays},
  \bibinfo{person}{Pietro Perona}, \bibinfo{person}{Deva Ramanan},
  \bibinfo{person}{Piotr Doll{'{a} }r}, {and} \bibinfo{person}{C.~Lawrence
  Zitnick}.} \bibinfo{year}{2014}\natexlab{}.
\newblock \showarticletitle{Microsoft {COCO:} Common Objects in Context}.
\newblock \bibinfo{journal}{\emph{CoRR}}  \bibinfo{volume}{abs/1405.0312}
  (\bibinfo{year}{2014}).
\newblock
\showeprint[arxiv]{1405.0312}
\urldef\tempurl%
\url{http://arxiv.org/abs/1405.0312}
\showURL{%
\tempurl}


\bibitem[Liu et~al\mbox{.}(2018)]%
        {liu2018picanet}
\bibfield{author}{\bibinfo{person}{Xueyang Liu}, \bibinfo{person}{Li Zhang},
  \bibinfo{person}{Lei Zhang}, \bibinfo{person}{Yuhao Zhang},
  \bibinfo{person}{Wen Ma}, {and} \bibinfo{person}{Tieniu Huang}.}
  \bibinfo{year}{2018}\natexlab{}.
\newblock \showarticletitle{PiCANet: Learning Pixel-wise Contextual Attention
  for Saliency Detection}. In \bibinfo{booktitle}{\emph{Proceedings of the
  European Conference on Computer Vision (ECCV)}}.
\newblock


\bibitem[Liu et~al\mbox{.}(2021)]%
        {liu2021swin}
\bibfield{author}{\bibinfo{person}{Ze Liu}, \bibinfo{person}{Yutong Lin},
  \bibinfo{person}{Yue Cao}, \bibinfo{person}{Han Hu}, \bibinfo{person}{Yixuan
  Wei}, \bibinfo{person}{Zheng Zhang}, \bibinfo{person}{Stephen Lin}, {and}
  \bibinfo{person}{Baining Guo}.} \bibinfo{year}{2021}\natexlab{}.
\newblock \showarticletitle{Swin transformer: Hierarchical vision transformer
  using shifted windows}. In \bibinfo{booktitle}{\emph{Proceedings of the
  IEEE/CVF international conference on computer vision}}.
  \bibinfo{pages}{10012--10022}.
\newblock


\bibitem[Lu et~al\mbox{.}(2021)]%
        {lu2021progressive}
\bibfield{author}{\bibinfo{person}{Yadong Lu}, \bibinfo{person}{Yinhao Zhu},
  \bibinfo{person}{Yang Yang}, \bibinfo{person}{Amir Said}, {and}
  \bibinfo{person}{Taco~S Cohen}.} \bibinfo{year}{2021}\natexlab{}.
\newblock \showarticletitle{Progressive neural image compression with nested
  quantization and latent ordering}. In \bibinfo{booktitle}{\emph{2021 IEEE
  International Conference on Image Processing (ICIP)}}. IEEE,
  \bibinfo{pages}{539--543}.
\newblock


\bibitem[Matsubara et~al\mbox{.}(2019)]%
        {matsubara2019distilled}
\bibfield{author}{\bibinfo{person}{Yoshitomo Matsubara}, \bibinfo{person}{Sabur
  Baidya}, \bibinfo{person}{Davide Callegaro}, \bibinfo{person}{Marco
  Levorato}, {and} \bibinfo{person}{Sameer Singh}.}
  \bibinfo{year}{2019}\natexlab{}.
\newblock \showarticletitle{Distilled split deep neural networks for
  edge-assisted real-time systems}. In \bibinfo{booktitle}{\emph{Proceedings of
  the 2019 Workshop on Hot Topics in Video Analytics and Intelligent Edges}}.
  \bibinfo{pages}{21--26}.
\newblock


\bibitem[Matsubara et~al\mbox{.}(2020)]%
        {matsubara2020head}
\bibfield{author}{\bibinfo{person}{Yoshitomo Matsubara},
  \bibinfo{person}{Davide Callegaro}, \bibinfo{person}{Sabur Baidya},
  \bibinfo{person}{Marco Levorato}, {and} \bibinfo{person}{Sameer Singh}.}
  \bibinfo{year}{2020}\natexlab{}.
\newblock \showarticletitle{Head network distillation: Splitting distilled deep
  neural networks for resource-constrained edge computing systems}.
\newblock \bibinfo{journal}{\emph{IEEE Access}}  \bibinfo{volume}{8}
  (\bibinfo{year}{2020}), \bibinfo{pages}{212177--212193}.
\newblock


\bibitem[Matsubara and Levorato(2020)]%
        {matsubara2020split}
\bibfield{author}{\bibinfo{person}{Yoshitomo Matsubara} {and}
  \bibinfo{person}{Marco Levorato}.} \bibinfo{year}{2020}\natexlab{}.
\newblock \showarticletitle{Split computing for complex object detectors:
  Challenges and preliminary results}.
\newblock \bibinfo{journal}{\emph{arXiv preprint arXiv:2007.13312}}
  (\bibinfo{year}{2020}).
\newblock


\bibitem[Matsubara and Levorato(2021)]%
        {matsubara2021neural}
\bibfield{author}{\bibinfo{person}{Yoshitomo Matsubara} {and}
  \bibinfo{person}{Marco Levorato}.} \bibinfo{year}{2021}\natexlab{}.
\newblock \showarticletitle{Neural compression and filtering for edge-assisted
  real-time object detection in challenged networks}. In
  \bibinfo{booktitle}{\emph{2020 25th International Conference on Pattern
  Recognition (ICPR)}}. IEEE, \bibinfo{pages}{2272--2279}.
\newblock


\bibitem[Matsubara et~al\mbox{.}(2022a)]%
        {matsubara2022split}
\bibfield{author}{\bibinfo{person}{Yoshitomo Matsubara}, \bibinfo{person}{Marco
  Levorato}, {and} \bibinfo{person}{Francesco Restuccia}.}
  \bibinfo{year}{2022}\natexlab{a}.
\newblock \showarticletitle{Split computing and early exiting for deep learning
  applications: Survey and research challenges}.
\newblock \bibinfo{journal}{\emph{Comput. Surveys}} \bibinfo{volume}{55},
  \bibinfo{number}{5} (\bibinfo{year}{2022}), \bibinfo{pages}{1--30}.
\newblock


\bibitem[Matsubara et~al\mbox{.}(2022b)]%
        {matsubara2022sc2}
\bibfield{author}{\bibinfo{person}{Yoshitomo Matsubara},
  \bibinfo{person}{Ruihan Yang}, \bibinfo{person}{Marco Levorato}, {and}
  \bibinfo{person}{Stephan Mandt}.} \bibinfo{year}{2022}\natexlab{b}.
\newblock \showarticletitle{SC2: Supervised compression for split computing}.
\newblock \bibinfo{journal}{\emph{arXiv preprint arXiv:2203.08875}}
  (\bibinfo{year}{2022}).
\newblock


\bibitem[Matsubara et~al\mbox{.}(2022c)]%
        {matsubara2022supervised}
\bibfield{author}{\bibinfo{person}{Yoshitomo Matsubara},
  \bibinfo{person}{Ruihan Yang}, \bibinfo{person}{Marco Levorato}, {and}
  \bibinfo{person}{Stephan Mandt}.} \bibinfo{year}{2022}\natexlab{c}.
\newblock \showarticletitle{Supervised compression for resource-constrained
  edge computing systems}. In \bibinfo{booktitle}{\emph{Proceedings of the
  IEEE/CVF Winter Conference on Applications of Computer Vision}}.
  \bibinfo{pages}{2685--2695}.
\newblock


\bibitem[Minnen et~al\mbox{.}(2018)]%
        {minnen2018joint}
\bibfield{author}{\bibinfo{person}{David Minnen}, \bibinfo{person}{Johannes
  Ball{\'e}}, {and} \bibinfo{person}{George~D Toderici}.}
  \bibinfo{year}{2018}\natexlab{}.
\newblock \showarticletitle{Joint autoregressive and hierarchical priors for
  learned image compression}.
\newblock \bibinfo{journal}{\emph{Advances in neural information processing
  systems}}  \bibinfo{volume}{31} (\bibinfo{year}{2018}).
\newblock


\bibitem[Pagliari et~al\mbox{.}(2020)]%
        {pagliari2020crime}
\bibfield{author}{\bibinfo{person}{Daniele~Jahier Pagliari},
  \bibinfo{person}{Roberta Chiaro}, \bibinfo{person}{Enrico Macii}, {and}
  \bibinfo{person}{Massimo Poncino}.} \bibinfo{year}{2020}\natexlab{}.
\newblock \showarticletitle{Crime: Input-dependent collaborative inference for
  recurrent neural networks}.
\newblock \bibinfo{journal}{\emph{IEEE Trans. Comput.}} \bibinfo{volume}{70},
  \bibinfo{number}{10} (\bibinfo{year}{2020}), \bibinfo{pages}{1626--1639}.
\newblock


\bibitem[Qin et~al\mbox{.}(2021)]%
        {qin2021boundary}
\bibfield{author}{\bibinfo{person}{Xuebin Qin}, \bibinfo{person}{Deng-Ping
  Fan}, \bibinfo{person}{Chenyang Huang}, \bibinfo{person}{Cyril Diagne},
  \bibinfo{person}{Zichen Zhang}, \bibinfo{person}{Adri{\`a}~Cabeza Sant'Anna},
  \bibinfo{person}{Albert Suarez}, \bibinfo{person}{Martin Jagersand}, {and}
  \bibinfo{person}{Ling Shao}.} \bibinfo{year}{2021}\natexlab{}.
\newblock \showarticletitle{Boundary-aware segmentation network for mobile and
  web applications}.
\newblock \bibinfo{journal}{\emph{arXiv preprint arXiv:2101.04704}}
  (\bibinfo{year}{2021}).
\newblock


\bibitem[Ramachandran and Sangaiah(2021)]%
        {ramachandran2021review}
\bibfield{author}{\bibinfo{person}{Anitha Ramachandran} {and}
  \bibinfo{person}{Arun~Kumar Sangaiah}.} \bibinfo{year}{2021}\natexlab{}.
\newblock \showarticletitle{A review on object detection in unmanned aerial
  vehicle surveillance}.
\newblock \bibinfo{journal}{\emph{International Journal of Cognitive Computing
  in Engineering}}  \bibinfo{volume}{2} (\bibinfo{year}{2021}),
  \bibinfo{pages}{215--228}.
\newblock


\bibitem[Redmon et~al\mbox{.}(2016a)]%
        {redmon2016yolo}
\bibfield{author}{\bibinfo{person}{Joseph Redmon}, \bibinfo{person}{Santosh
  Divvala}, \bibinfo{person}{Ross Girshick}, {and} \bibinfo{person}{Ali
  Farhadi}.} \bibinfo{year}{2016}\natexlab{a}.
\newblock \showarticletitle{You Only Look Once: Unified, Real-Time Object
  Detection}. In \bibinfo{booktitle}{\emph{Proceedings of the IEEE Conference
  on Computer Vision and Pattern Recognition}}. \bibinfo{pages}{779--788}.
\newblock


\bibitem[Redmon et~al\mbox{.}(2016b)]%
        {redmon2016you}
\bibfield{author}{\bibinfo{person}{Joseph Redmon}, \bibinfo{person}{Santosh
  Divvala}, \bibinfo{person}{Ross Girshick}, {and} \bibinfo{person}{Ali
  Farhadi}.} \bibinfo{year}{2016}\natexlab{b}.
\newblock \showarticletitle{You Only Look Once: Unified, Real-Time Object
  Detection}.
\newblock \bibinfo{journal}{\emph{Proceedings of the IEEE Conference on
  Computer Vision and Pattern Recognition}} (\bibinfo{year}{2016}),
  \bibinfo{pages}{779--788}.
\newblock


\bibitem[Ren et~al\mbox{.}(2017)]%
        {ren2015faster}
\bibfield{author}{\bibinfo{person}{Shaoqing Ren}, \bibinfo{person}{Kaiming He},
  \bibinfo{person}{Ross Girshick}, {and} \bibinfo{person}{Jian Sun}.}
  \bibinfo{year}{2017}\natexlab{}.
\newblock \showarticletitle{Faster {R-CNN}: Towards Real-Time Object Detection
  with Region Proposal Networks}.
\newblock \bibinfo{journal}{\emph{IEEE Transactions on Pattern Analysis and
  Machine Intelligence}} \bibinfo{volume}{39}, \bibinfo{number}{6}
  (\bibinfo{year}{2017}), \bibinfo{pages}{1137--1149}.
\newblock


\bibitem[Said and Pearlman(1996)]%
        {said1996new}
\bibfield{author}{\bibinfo{person}{Amir Said} {and} \bibinfo{person}{William~A
  Pearlman}.} \bibinfo{year}{1996}\natexlab{}.
\newblock \showarticletitle{A new, fast, and efficient image codec based on set
  partitioning in hierarchical trees}.
\newblock \bibinfo{journal}{\emph{IEEE Transactions on circuits and systems for
  video technology}} \bibinfo{volume}{6}, \bibinfo{number}{3}
  (\bibinfo{year}{1996}), \bibinfo{pages}{243--250}.
\newblock


\bibitem[Selvaraju et~al\mbox{.}(2020)]%
        {selvaraju2017gradcam}
\bibfield{author}{\bibinfo{person}{Ramprasaath~R. Selvaraju},
  \bibinfo{person}{Michael Cogswell}, \bibinfo{person}{Abhishek Das},
  \bibinfo{person}{Ramakrishna Vedantam}, \bibinfo{person}{Devi Parikh}, {and}
  \bibinfo{person}{Dhruv Batra}.} \bibinfo{year}{2020}\natexlab{}.
\newblock \showarticletitle{Grad-CAM: Visual Explanations from Deep Networks
  via Gradient-based Localization}.
\newblock \bibinfo{journal}{\emph{International Journal of Computer Vision}}
  \bibinfo{volume}{128}, \bibinfo{number}{2} (\bibinfo{year}{2020}),
  \bibinfo{pages}{336--359}.
\newblock


\bibitem[Shao and Zhang(2020)]%
        {shao2020bottlenet++}
\bibfield{author}{\bibinfo{person}{Jiawei Shao} {and} \bibinfo{person}{Jun
  Zhang}.} \bibinfo{year}{2020}\natexlab{}.
\newblock \showarticletitle{Bottlenet++: An end-to-end approach for feature
  compression in device-edge co-inference systems}. In
  \bibinfo{booktitle}{\emph{2020 IEEE International Conference on
  Communications Workshops (ICC Workshops)}}. IEEE, \bibinfo{pages}{1--6}.
\newblock


\bibitem[Sikora et~al\mbox{.}(2019)]%
        {sikora2019test}
\bibfield{author}{\bibinfo{person}{Axel Sikora}, \bibinfo{person}{Manuel
  Schappacher}, \bibinfo{person}{Zubair Amjad}, {et~al\mbox{.}}}
  \bibinfo{year}{2019}\natexlab{}.
\newblock \showarticletitle{Test and measurement of LPWAN and cellular IoT
  networks in a unified testbed}. In \bibinfo{booktitle}{\emph{2019 IEEE 17th
  International Conference on Industrial Informatics (INDIN)}},
  Vol.~\bibinfo{volume}{1}. IEEE, \bibinfo{pages}{1521--1527}.
\newblock


\bibitem[Skodras et~al\mbox{.}(2001)]%
        {skodras2001jpeg}
\bibfield{author}{\bibinfo{person}{Athanassios Skodras},
  \bibinfo{person}{Charilaos Christopoulos}, {and} \bibinfo{person}{Touradj
  Ebrahimi}.} \bibinfo{year}{2001}\natexlab{}.
\newblock \showarticletitle{The JPEG 2000 still image compression standard}.
\newblock \bibinfo{journal}{\emph{IEEE Signal processing magazine}}
  \bibinfo{volume}{18}, \bibinfo{number}{5} (\bibinfo{year}{2001}),
  \bibinfo{pages}{36--58}.
\newblock


\bibitem[Su et~al\mbox{.}(2020)]%
        {su2020scalable}
\bibfield{author}{\bibinfo{person}{Rige Su}, \bibinfo{person}{Zhengxue Cheng},
  \bibinfo{person}{Heming Sun}, {and} \bibinfo{person}{Jiro Katto}.}
  \bibinfo{year}{2020}\natexlab{}.
\newblock \showarticletitle{Scalable learned image compression with a recurrent
  neural networks-based hyperprior}. In \bibinfo{booktitle}{\emph{2020 IEEE
  International Conference on Image Processing (ICIP)}}. IEEE,
  \bibinfo{pages}{3369--3373}.
\newblock


\bibitem[Tan and Le(2019)]%
        {tan2019efficientnet}
\bibfield{author}{\bibinfo{person}{Mingxing Tan} {and} \bibinfo{person}{Quoc
  Le}.} \bibinfo{year}{2019}\natexlab{}.
\newblock \showarticletitle{Efficientnet: Rethinking model scaling for
  convolutional neural networks}. In \bibinfo{booktitle}{\emph{International
  conference on machine learning}}. PMLR, \bibinfo{pages}{6105--6114}.
\newblock


\bibitem[Toderici et~al\mbox{.}(2015)]%
        {toderici2015variable}
\bibfield{author}{\bibinfo{person}{George Toderici}, \bibinfo{person}{Sean~M
  O'Malley}, \bibinfo{person}{Sung~Jin Hwang}, \bibinfo{person}{Damien
  Vincent}, \bibinfo{person}{David Minnen}, \bibinfo{person}{Shumeet Baluja},
  \bibinfo{person}{Michele Covell}, {and} \bibinfo{person}{Rahul Sukthankar}.}
  \bibinfo{year}{2015}\natexlab{}.
\newblock \showarticletitle{Variable rate image compression with recurrent
  neural networks}.
\newblock \bibinfo{journal}{\emph{arXiv preprint arXiv:1511.06085}}
  (\bibinfo{year}{2015}).
\newblock


\bibitem[Toderici et~al\mbox{.}(2017)]%
        {toderici2017full}
\bibfield{author}{\bibinfo{person}{George Toderici}, \bibinfo{person}{Damien
  Vincent}, \bibinfo{person}{Nick Johnston}, \bibinfo{person}{Sung Jin~Hwang},
  \bibinfo{person}{David Minnen}, \bibinfo{person}{Joel Shor}, {and}
  \bibinfo{person}{Michele Covell}.} \bibinfo{year}{2017}\natexlab{}.
\newblock \showarticletitle{Full resolution image compression with recurrent
  neural networks}. In \bibinfo{booktitle}{\emph{Proceedings of the IEEE
  conference on Computer Vision and Pattern Recognition}}.
  \bibinfo{pages}{5306--5314}.
\newblock


\bibitem[Uijlings et~al\mbox{.}(2013)]%
        {uijlings2013selective}
\bibfield{author}{\bibinfo{person}{Jasper~RR Uijlings},
  \bibinfo{person}{Koen~EA van~de Sande}, \bibinfo{person}{Theo Gevers}, {and}
  \bibinfo{person}{Arnold~WM Smeulders}.} \bibinfo{year}{2013}\natexlab{}.
\newblock \showarticletitle{Selective Search for Object Recognition}. In
  \bibinfo{booktitle}{\emph{International Journal of Computer Vision}},
  Vol.~\bibinfo{volume}{104}. Springer, \bibinfo{pages}{154--171}.
\newblock


\bibitem[Vijayan et~al\mbox{.}(2022)]%
        {vijayan2022video}
\bibfield{author}{\bibinfo{person}{Alpha Vijayan}, \bibinfo{person}{B
  Meenaskshi}, \bibinfo{person}{Aditya Pandey}, \bibinfo{person}{Akshat Patel},
  {and} \bibinfo{person}{Arohi Jain}.} \bibinfo{year}{2022}\natexlab{}.
\newblock \showarticletitle{Video anomaly detection in surveillance cameras}.
  In \bibinfo{booktitle}{\emph{2022 International Conference for Advancement in
  Technology (ICONAT)}}. IEEE, \bibinfo{pages}{1--4}.
\newblock


\bibitem[Wallace(1991)]%
        {wallace1991jpeg}
\bibfield{author}{\bibinfo{person}{Gregory~K Wallace}.}
  \bibinfo{year}{1991}\natexlab{}.
\newblock \showarticletitle{The JPEG still picture compression standard}.
\newblock \bibinfo{journal}{\emph{Commun. ACM}} \bibinfo{volume}{34},
  \bibinfo{number}{4} (\bibinfo{year}{1991}), \bibinfo{pages}{30--44}.
\newblock


\bibitem[Wallace(1992)]%
        {wallace1992jpeg}
\bibfield{author}{\bibinfo{person}{Gregory~K Wallace}.}
  \bibinfo{year}{1992}\natexlab{}.
\newblock \showarticletitle{The JPEG still picture compression standard}.
\newblock \bibinfo{journal}{\emph{IEEE transactions on consumer electronics}}
  \bibinfo{volume}{38}, \bibinfo{number}{1} (\bibinfo{year}{1992}),
  \bibinfo{pages}{xviii--xxxiv}.
\newblock


\bibitem[Wang et~al\mbox{.}(2021)]%
        {wang2020scorecam}
\bibfield{author}{\bibinfo{person}{Haofan Wang}, \bibinfo{person}{Zifan Wang},
  \bibinfo{person}{Mengnan Du}, \bibinfo{person}{Fan Yang},
  \bibinfo{person}{Zijian Zhang}, \bibinfo{person}{Sirui Ding},
  \bibinfo{person}{Piotr Mardziel}, \bibinfo{person}{Xia Hu}, {and}
  \bibinfo{person}{Xia Hu}.} \bibinfo{year}{2021}\natexlab{}.
\newblock \showarticletitle{Score-CAM: Score-Weighted Visual Explanations for
  Convolutional Neural Networks}.
\newblock \bibinfo{journal}{\emph{IEEE Transactions on Neural Networks and
  Learning Systems}} \bibinfo{volume}{32}, \bibinfo{number}{11}
  (\bibinfo{year}{2021}), \bibinfo{pages}{5185--5198}.
\newblock


\bibitem[Wang et~al\mbox{.}(2019)]%
        {wang2019bilateral}
\bibfield{author}{\bibinfo{person}{Lijun Wang}, \bibinfo{person}{Huchuan Lu},
  \bibinfo{person}{Yifan Wang}, \bibinfo{person}{Mengyang Feng},
  \bibinfo{person}{Dong Wang}, {and} \bibinfo{person}{Baocai Yin}.}
  \bibinfo{year}{2019}\natexlab{}.
\newblock \showarticletitle{Bilateral Multi-Pooling with Bi-Directional Message
  Passing for Salient Object Detection}. In
  \bibinfo{booktitle}{\emph{Proceedings of the IEEE Conference on Computer
  Vision and Pattern Recognition (CVPR)}}.
\newblock


\bibitem[Wang et~al\mbox{.}(2018)]%
        {wang2018recurrent}
\bibfield{author}{\bibinfo{person}{Xinyu Wang}, \bibinfo{person}{Xinyuan Cao},
  \bibinfo{person}{Chunhua Shen}, {and} \bibinfo{person}{Liang Lin}.}
  \bibinfo{year}{2018}\natexlab{}.
\newblock \showarticletitle{Recurrent Residual Module for Fast Inference in
  Videos}. In \bibinfo{booktitle}{\emph{Proceedings of the IEEE Conference on
  Computer Vision and Pattern Recognition (CVPR)}}.
\newblock


\bibitem[WebP({[n.\,d.]})]%
        {webp}
WebP \bibinfo{year}{[n.\,d.]}\natexlab{}.
\newblock \bibinfo{title}{WebP}.
\newblock
  \bibinfo{howpublished}{\url{https://developers.google.com/speed/webp/docs/compression}}.
\newblock
\newblock
\shownote{Accessed: 2023-02-14}.


\bibitem[Yao et~al\mbox{.}(2020)]%
        {yao2020deep}
\bibfield{author}{\bibinfo{person}{Shuochao Yao}, \bibinfo{person}{Jinyang Li},
  \bibinfo{person}{Dongxin Liu}, \bibinfo{person}{Tianshi Wang},
  \bibinfo{person}{Shengzhong Liu}, \bibinfo{person}{Huajie Shao}, {and}
  \bibinfo{person}{Tarek Abdelzaher}.} \bibinfo{year}{2020}\natexlab{}.
\newblock \showarticletitle{Deep compressive offloading: Speeding up neural
  network inference by trading edge computation for network latency}. In
  \bibinfo{booktitle}{\emph{Proceedings of the 18th Conference on Embedded
  Networked Sensor Systems}}. \bibinfo{pages}{476--488}.
\newblock


\bibitem[Ye et~al\mbox{.}(2023)]%
        {ye2023accelir}
\bibfield{author}{\bibinfo{person}{Juncheol Ye}, \bibinfo{person}{Hyunho Yeo},
  \bibinfo{person}{Jinwoo Park}, {and} \bibinfo{person}{Dongsu Han}.}
  \bibinfo{year}{2023}\natexlab{}.
\newblock \showarticletitle{AccelIR: Task-Aware Image Compression for
  Accelerating Neural Restoration}. In \bibinfo{booktitle}{\emph{Proceedings of
  the IEEE/CVF Conference on Computer Vision and Pattern Recognition}}.
  \bibinfo{pages}{18216--18226}.
\newblock


\bibitem[Zainab et~al\mbox{.}(2023)]%
        {zainab2023lighteq}
\bibfield{author}{\bibinfo{person}{Tayyaba Zainab}, \bibinfo{person}{Jens
  Karstens}, {and} \bibinfo{person}{Olaf Landsiedel}.}
  \bibinfo{year}{2023}\natexlab{}.
\newblock \showarticletitle{LightEQ: On-Device Earthquake Detection with
  Embedded Machine Learning}. In \bibinfo{booktitle}{\emph{Proceedings of the
  8th ACM/IEEE Conference on Internet of Things Design and Implementation}}.
  \bibinfo{pages}{130--143}.
\newblock


\bibitem[Zeng et~al\mbox{.}(2019)]%
        {zeng2019boomerang}
\bibfield{author}{\bibinfo{person}{Liekang Zeng}, \bibinfo{person}{En Li},
  \bibinfo{person}{Zhi Zhou}, {and} \bibinfo{person}{Xu Chen}.}
  \bibinfo{year}{2019}\natexlab{}.
\newblock \showarticletitle{Boomerang: On-demand cooperative deep neural
  network inference for edge intelligence on the industrial Internet of
  Things}.
\newblock \bibinfo{journal}{\emph{IEEE Network}} \bibinfo{volume}{33},
  \bibinfo{number}{5} (\bibinfo{year}{2019}), \bibinfo{pages}{96--103}.
\newblock


\bibitem[Zhang et~al\mbox{.}(2018)]%
        {zhang2018nested}
\bibfield{author}{\bibinfo{person}{Dong Zhang}, \bibinfo{person}{Yuchen Han},
  \bibinfo{person}{Ming-Hsuan Yang}, \bibinfo{person}{Tong Zhang},
  \bibinfo{person}{Wei Liu}, \bibinfo{person}{Xiaokang Yang}, {and}
  \bibinfo{person}{Jing Guo}.} \bibinfo{year}{2018}\natexlab{}.
\newblock \showarticletitle{Nested Lateral Descriptive Features for Salient
  Object Detection}. In \bibinfo{booktitle}{\emph{Proceedings of the IEEE
  Conference on Computer Vision and Pattern Recognition (CVPR)}}.
\newblock


\end{thebibliography}

\end{document}